%% file: main.tex
\documentclass[runningheads]{llncs}

\usepackage{eccv}

\usepackage{eccvabbrv}
\usepackage{colortbl} 
\usepackage{xcolor} 
\usepackage{booktabs} 
\usepackage{caption}  
\usepackage{amssymb}
\usepackage{bm}
\usepackage{multicol}
\usepackage{multirow}
\usepackage{subcaption}
\usepackage{amsmath}
\usepackage{makecell}
\usepackage{amsfonts}
\usepackage{pifont}
\definecolor{ForestGreen}{RGB}{34,139,34}
\definecolor{BrickRed}{RGB}{182,49,44}
\newcommand{\cmark}{\textcolor{ForestGreen}{\ding{51}}} 
\newcommand{\xmark}{\textcolor{BrickRed}{\ding{55}}}    

\usepackage{wrapfig}
\usepackage{amssymb} 

\usepackage{graphicx}
\usepackage{booktabs}

\usepackage[accsupp]{axessibility}  

\usepackage[pagebackref,breaklinks,colorlinks,citecolor=eccvblue]{hyperref}

\usepackage{orcidlink}

\begin{document}

\title{Neural Metamorphosis} 

\titlerunning{NeuMeta}

\author{Xingyi Yang\orcidlink{0000-1111-2222-3333} \and
Xinchao Wang\thanks{Corresponding author.}\orcidlink{0000-0003-0057-1404}}

\authorrunning{X. Yang et al.}

\institute{National University of Singapore \\
\email{xyang@u.nus.edu}, \email{xinchao@nus.edu.sg}\\
\href{https://adamdad.github.io/neumeta/}{\url{https://adamdad.github.io/neumeta/}}}

\maketitle

\input{sec/0_abstract}    
\input{sec/1_intro}

\input{sec/1_1_related}

\input{sec/4_method}

\input{sec/5_exp}

\input{sec/6_conclusion}

\section*{Acknowledgement}
This project is supported by
the National Research Foundation, Singapore, under its AI Singapore Programme (AISG Award No: AISG2-RP-2021-023).
%
%
\bibliographystyle{splncs04}
\bibliography{main}
\end{document}


\clearpage
\setcounter{page}{1}
\maketitlesupplementary

This supplementary material delves into ``Neural Metamorphosis'', starting with the its pipeline and pseudo-code in Section~\ref{sec:pipeline}. Section~\ref{sec:proof} presents a proof of the orthogonal property of total variation. Section~\ref{sec:generalized} establishes the INR as a generalized form of current continuous function representation. In Section~\ref{sec:depth}, we extend our experiments to network depth morphing. Section~\ref{sec:permute} compares our weight permutation strategy with existing methods. Ablation studies in Section~\ref{sec:ablation} examine the impact of model architecture, block-based INR, and EMA. Section~\ref{sec:comparenern} contrasts \texttt{NeuMeta} with similar works, supported by visual analyses in Section~\ref{sec:vis}. Later we present further experimental details. The document concludes with a discussion on scalability and limitations, suggesting directions for future research.

\section{Pipeline for Neural Metamorphosis}
\label{sec:pipeline}
In this section, we describe the pipeline for the Neural Metamorphosis, which is composed of three main stages: (1) Weight Permutation (2) INR Training (3) Weight Sampling. The pseudo-code is outlined in Algorithm~\ref{algorithm:neumeta_permute},~\ref{algorithm:neumeta_train} and~\ref{algorithm:neumeta_sample}.

\begin{itemize}
    \item \textbf{Weight Permutation.} This step involves modifying and smoothing a trained neural network's weights \(\mathbf{W}\) by applying an optimal permutation matrix \(P^*\) on each clique graph. The aim is to generate a new, smoother set of weights, \(\mathbf{W}^{(\text{smooth})}\), which are more readily learnable by the INR.
    \item \textbf{INR Training.} Leveraging the smoothed weights and the dataset, this step develops an Implicit Neural Network \(F(\cdot;\theta)\). The primary objective here is to iteratively update the network's weights to optimize the INR. This optimization aims to minimize the overall loss on the dataset, thereby refining the performance of the network.
    \item \textbf{Weight Sampling.} The final stage is centered on extracting  weights for the target network architecture \(\mathbf{i}\) from INR.  This process includes collecting $K$ samples and averaging their weights to create a customized weight matrix $\mathbf{W}.$ This matrix is specifically designed to suit the chosen architecture, ensuring that the target network is optimally configured for its intended tasks.
\end{itemize}

\begin{algorithm}[tbh]
\caption{Neural Metamorphosis -- Weight Permutation}
\vspace{3mm}
\label{algorithm:neumeta_permute}
\begin{algorithmic}[1]
\Input Trained neural network $f(\cdot;\mathbf{W})$ with dependency graph $G=(V,E)$. The weight of $i$-th layer is denoted as $\mathbf{W}_i$.
\Output The permuted and smoothed weight $\mathbf{W}^{(\text{smooth})}$
\For{$C=(V_C,E_C)$ \textbf{such that} ($C\subset G$ and $C$ is a clique)}
\State Solve for optimal permutation matrix $P^*$
\begin{align*}
    P^* &= \operatorname*{argmin}_{P} \sum_{e_{ij} \in E_C} \Big(TV_{\text{out}}(P\mathbf{W}_i) + TV_{\text{in}}(\mathbf{W}_j P^{-1})\Big)
\end{align*}
\For{$e_{ij}$ \textbf{such that} $e_{ij} \in E_C$}
\State Permute the weights according to the $P*$
\begin{align*}
    \mathbf{W}^{(\text{smooth})}_i \leftarrow P\mathbf{W}_i; \mathbf{W}^{(\text{smooth})}_j \leftarrow \mathbf{W}_i P^{-1}
\end{align*}
 \EndFor
  \EndFor
\State \textbf{return} $\mathbf{W}^{(\text{smooth})}$.
\end{algorithmic}
\vspace{3mm}
\end{algorithm}


\begin{algorithm}[tbh]
\caption{Neural Metamorphosis -- INR Training}
\vspace{3mm}
\label{algorithm:neumeta_train}
\begin{algorithmic}[1]
\Input A trained neural network $f(\cdot;\mathbf{W}^{(\text{smooth})})$, a training set  $D_{tr}=\{\mathbf{x}_i, y_i\}_{i=1}^M$ and predefined configuration pool $\{\mathbf{i}_k\}_{k=1}^K$, total train iteration $T$.
\Output Implicit neural network $F(\cdot;\theta)$
\For{$t$ \textbf{to} $T$}
\State Sample a config from pool $\mathbf{i} \in \{\mathbf{i}_k\}_{k=1}^K$.
\For{$\mathbf{j}$ \textbf{such that} $\mathbf{j} \in \mathcal{J}_{\mathbf{i}}$}
\State $(\mathbf{i}',\mathbf{j}') \leftarrow (\mathbf{i},\mathbf{j}) + \bm\epsilon$ where $\bm\epsilon\sim \text{U}(-\mathbf{a}, \mathbf{a})$.
\State $\mathbf{v}' \leftarrow \left[\frac{l'}{L'}, \frac{c'_{\text{in}}}{C'_{\text{in}}}, \frac{c_'{\text{out}}}{C'_{\text{out}}}, \frac{L'}{N}, \frac{C'_{\text{in}}}{N},  \frac{C'_{\text{out}}}{N}\right]$.
\State Generate weight $w_{(\mathbf{i},\mathbf{j})} \leftarrow F(\gamma_{\text{PE}}(\mathbf{v}');\theta)$.
\EndFor
\State Sample a batch of train data $(X,Y)$.
\State Inference with the generated weights $\hat{Y} = f(X;\mathbf{W})$.
\State Calculate total loss $\mathcal{L}= \mathcal{L}_{\text{task}} + \lambda_1 \mathcal{L}_{\text{recon}} + \lambda_2 \mathcal{L}_{\text{reg}}$.
\State Calculate gradient$\nabla_{\theta} \mathcal{L} = \frac{\partial \mathcal{L}_{\text{task}}}{\partial W} \frac{\partial W}{\partial \theta} + \lambda_1 \frac{\partial \mathcal{L}_{\text{recon}}}{\partial \theta} + \lambda_2\frac{\partial \mathcal{L}_{\text{reg}}}{\partial \theta}$.
\State Update the INR's weight $\theta \leftarrow \theta - \alpha \nabla_{\theta} \mathcal{L}$
\EndFor
\State \textbf{return} $F(\cdot;\theta)$.
\end{algorithmic}
\vspace{3mm}
\end{algorithm}

\begin{algorithm}[tbh]
\caption{Neural Metamorphosis -- Weight Sampling}
\label{algorithm:neumeta_sample}
\begin{algorithmic}[1]
\Input Trained INR $F(\cdot;\theta)$, number of sampling $K$ and desired architecture $\mathbf{i}$.
\Output Weight $\mathbf{W}$ corresponds to $\mathbf{i}$.
\State Initialize all elements in $\mathbf{W}$ to be 0.
\For{$k=1$ \textbf{to} $K$}
\State $(\mathbf{i}',\mathbf{j}') \leftarrow (\mathbf{i},\mathbf{j}) + \bm\epsilon$ where $\bm\epsilon\sim \text{U}(-\mathbf{a}, \mathbf{a})$.
\State $\mathbf{v}' \leftarrow \left[\frac{l'}{L'}, \frac{c'_{\text{in}}}{C'_{\text{in}}}, \frac{c_'{\text{out}}}{C'_{\text{out}}}, \frac{L'}{N}, \frac{C'_{\text{in}}}{N},  \frac{C'_{\text{out}}}{N}\right]$.
\State Generate and average $w_{(\mathbf{i},\mathbf{j})} \leftarrow  w_{(\mathbf{i},\mathbf{j})} + \frac{1}{K}F(\gamma_{\text{PE}}(\mathbf{v}');\theta)$.
\EndFor
\State \textbf{return} $\mathbf{W}$.
\end{algorithmic}
\end{algorithm}

\section{Orthogonality of Total Variation: A Proof}
\label{sec:proof}
\textbf{Definitions and Notations:}
\begin{itemize}
    \item Let \( \mathbf{W} \) be an \( m \times n \) matrix.
    \item \( TV_{\text{in}}(\mathbf{W}) \) denotes the sum of the absolute differences between consecutive elements within each row of \( \mathbf{W} \).
    \begin{equation}
    TV_{\text{in}}(\mathbf{W}) = \sum_{i=1}^{m} \sum_{j=1}^{n-1} \left| w_{i,j+1} - w_{i,j} \right|
    \end{equation}
    \item \( TV_{\text{out}}(\mathbf{W}) \) denotes the sum of the absolute differences between consecutive rows (i.e., row-wise TV). For an \( m \times n \) matrix, it's defined as:
    \begin{equation}
    TV_{\text{out}}(\mathbf{W}) = \sum_{i=1}^{m-1} \sum_{j=1}^{n} \left| w_{i+1,j} - w_{i,j} \right|
    \end{equation}
\end{itemize}

With these notations, the total variation of \( \mathbf{W} \) can be expressed as:
\[
TV(\mathbf{W}) = TV_{\text{in}}(\mathbf{W}) + TV_{\text{out}}(\mathbf{W})
\]

\textbf{Proof:}
When considering a permutation matrix 
$P$, we observe its effects on matrix 
$\mathbf{W}$ in terms of column and row permutations.



\textbf{Case 1: Permutation of columns (\( P\mathbf{W} \))}

Column permutation, executed by 
\( P \), rearranges the columns of 
$\mathbf{W}$ but does not affect the relative differences within each row. Consequently, the total variation within rows, 
\( TV_{\text{in}} \), remains constant:
\begin{align}
&TV_{\text{in}}(P\mathbf{W}) = TV_{\text{in}}(\mathbf{W}) \quad \text{(row TV are unchanged)} 
\end{align}

\textbf{Case 2: Permutation of rows (\( \mathbf{W}P \))}

Conversely, row permutation modifies the order of rows in 
$\mathbf{W}$ without altering the internal composition of each row. Thus, the total variation between rows, 
$TV_{\text{out}}$, stays unchanged:
\begin{align}
&TV_{\text{out}}(\mathbf{W}P) = TV_{\text{out}}(\mathbf{W}) \quad \text{(column TV are unchanged)} 
\end{align}

From the aforementioned cases, we can clearly infer that a permutation applied to one dimension of the matrix does not influence the total variation in its orthogonal dimension.

This proof underscores the \emph{axis-orthogonal} characteristic of Total Variation (TV). However, it's crucial to note that this property may not extend to other forms of smoothness metrics. For example, in cases where smoothness is evaluated with respect to diagonal neighbors, defined by the expression $\sum_{i=1}^{m-1} \sum_{j=1}^{n-1} \left| w_{i+1,j+1} - w_{i,j} \right|$, the same orthogonal properties of TV may not be applicable.

\section{INR as Generalized Continuous Function}
\label{sec:generalized}
Currently, two methods exist for representing neural network weights as continuous functions: the piece-wise linear function approach~\cite{le2007continuous} and as using kernel method~\cite{Solodskikh_2023_CVPR}. This section aims to establish that the Implicit Neural Representation (INR) offers a more generalized approach compared to these traditional methods. 

We show the proof in 1D signal, which can be easily extended to higher-dimensional scenarios.

\subsection{INR Generalizes to Piece-wise Linear Method}
In this section, we demonstrate that an INR can be viewed as a generalized form of Piece-wise Linear Function (PLF). We establish this by showing that a Multi-Layer Perceptron (MLP) can approximate any PLF, particularly using a single-layer neural network with ReLU activation.

\noindent\textbf{Definition of Piece-wise Linear Function}

A Piece-wise Linear Function is defined as a function consisting of several linear segments. Formally, for a PLF $f(x)$ over an interval $[a, b]$, it is expressed as:

\[
f(x) = 
\begin{cases} 
a_1x + b_1 & \text{if } x \in [x_0, x_1], \\
a_2x + b_2 & \text{if } x \in (x_1, x_2], \\
\quad \vdots \\
a_nx + b_n & \text{if } x \in (x_{n-1}, x_n],
\end{cases}
\]
where \( x_0 = a \) and \( x_n = b \), and \( a_i, b_i \) are constants.

\noindent\textbf{Construction Using a Single-Layer Neural Network:} 

Consider a neural network with a single layer having \( n \) neurons, each with a ReLU activation function. The output of the \( i \)-th neuron for an input \( x \) is 
\[
y_i = \text{ReLU}(w_i x + b_i)= \max(0,w_i x + b_i),
\]
where \( w_i \) and \( b_i \) are the weight and bias of the \( i \)-th neuron, respectively.

To approximate the PLF, we align \( w_i = a_i \) and choose \( b_i \) so that the line \( y = w_i x + b_i \) aligns with the \( i \)-th segment of the PLF. As such, The final output of the neural network is exactly the same as the original PLF
\[
f_{NN}(x) = \sum_{i=1}^{n} y_i = \sum_{i=1}^{n} \text{ReLU}(a_i x + b_i).
\]

\noindent\textbf{Proof of Approximation:} \\
For any given PLF, we can find a set of weights and biases \( \{w_i, b_i\} \) for a single-layer neural network with ReLU activation such that its output \( f_{NN}(x) \) approximates the PLF \( f(x) \) over the interval \( [a, b] \). This leverages the property of ReLU to emulate piece-wise linear segments.

\subsection{INR Generalizes to Kernel Method}

To demonstrate that an INR generalizes to any kernel method, we show that kernel method can be implemented using a neural network with the kernel function as the activation function.

\noindent\textbf{Definition of Kernel Method} 

Kernel methods are a class of algorithms used in machine learning for pattern analysis. A kernel function takes two inputs and outputs a similarity measure. In mathematical terms, a kernel 
$K(x,y)$ is a function that for all 
$x,y\in\mathcal{X}$, where 
$\mathcal{X}$ is the input space, satisfies certain properties (e.g., symmetry, positive definiteness).

For example, In~\cite{Solodskikh_2023_CVPR}, they use a bicubic kernel 
$K(x,y)$  that computes the weighted average of points around a given position 
$y$ from 
$x$. The kernel typically employs cubic polynomials and is defined as:

\[
K(x, y) = {\tiny\begin{cases} 
(1.5 |x - y|^3 - 2.5 |x - y|^2 + 1) & \text{for } |x - y| < 1, \\
(-0.5 |x - y|^3 + 2.5 |x - y|^2 - 4 |x - y| + 2) & \text{for } 1 \leq |x - y| < 2, \\
0 & \text{otherwise}.
\end{cases}}
\]

\noindent\textbf{Construction Using Neural Network:} \\
In an INR framework, a neural network can utilize the kernel function as its last layer's activation function. For an input \( x \) and a set of data points \( \{y_i\}_{i=1}^N \), the network's output is formulated as:
\[
f_{NN}(x) = \sum_{i=1}^{N} w_i K(MLP(x), MLP(y_i)) + b,
\]
where \( w_i \) are the weights, \( b \) is the bias, and \( N \) is the number of data points.

\noindent\textbf{Proof of Approximation:} \\
The Universal Approximation Theorem implies that if the MLP is used to replace the learned sampling points as described in~\cite{Solodskikh_2023_CVPR}, the two formulations essentially become identical.  By adjusting the weights \( \{w_i\} \) and bias \( b \), the network can replicate traditional kernel methods.

In the preceding two subsections, we demonstrated that our INR serves as a generalized parameterization method for traditional continuous weight neural networks. This provides a strong theoretical foundation that reinforces the superior empirical results observed with our approach.

\section{Extension: Depth Morphing}
\label{sec:depth}
In this extension study, we explore the concept of \emph{morphing network depth} as opposed to network width done in the main paper. This advances our goal of morphing arbitrary network weights beyond just the channel number.

\noindent\textbf{Experiment Setup.} This experiment utilizes the MNIST dataset with a specially structured ResNet. This ResNet is constructed in two stages, each comprising a variable number of residual blocks, denoted as $L1$ for stage 1 and $L2$ for stage 2, respectively. Each residual block contains two $3\times3$ convolutional layers with ReLU activation. We initiate our process with a pretrained model where $L1=2$ and $L2=2$. We then train an INR by varying $L1$ and $L2$. During training, we sample $L1, L2 \in \{1,2,3\}$, but for testing, we evaluate performance across a wider range with $L1, L2 \in \{1,2,3,4,5,6,7,8,9\}$. In this experiment, we employ a single INR to fit all layers.

\begin{figure}
    \centering
    \includegraphics[width=\linewidth]{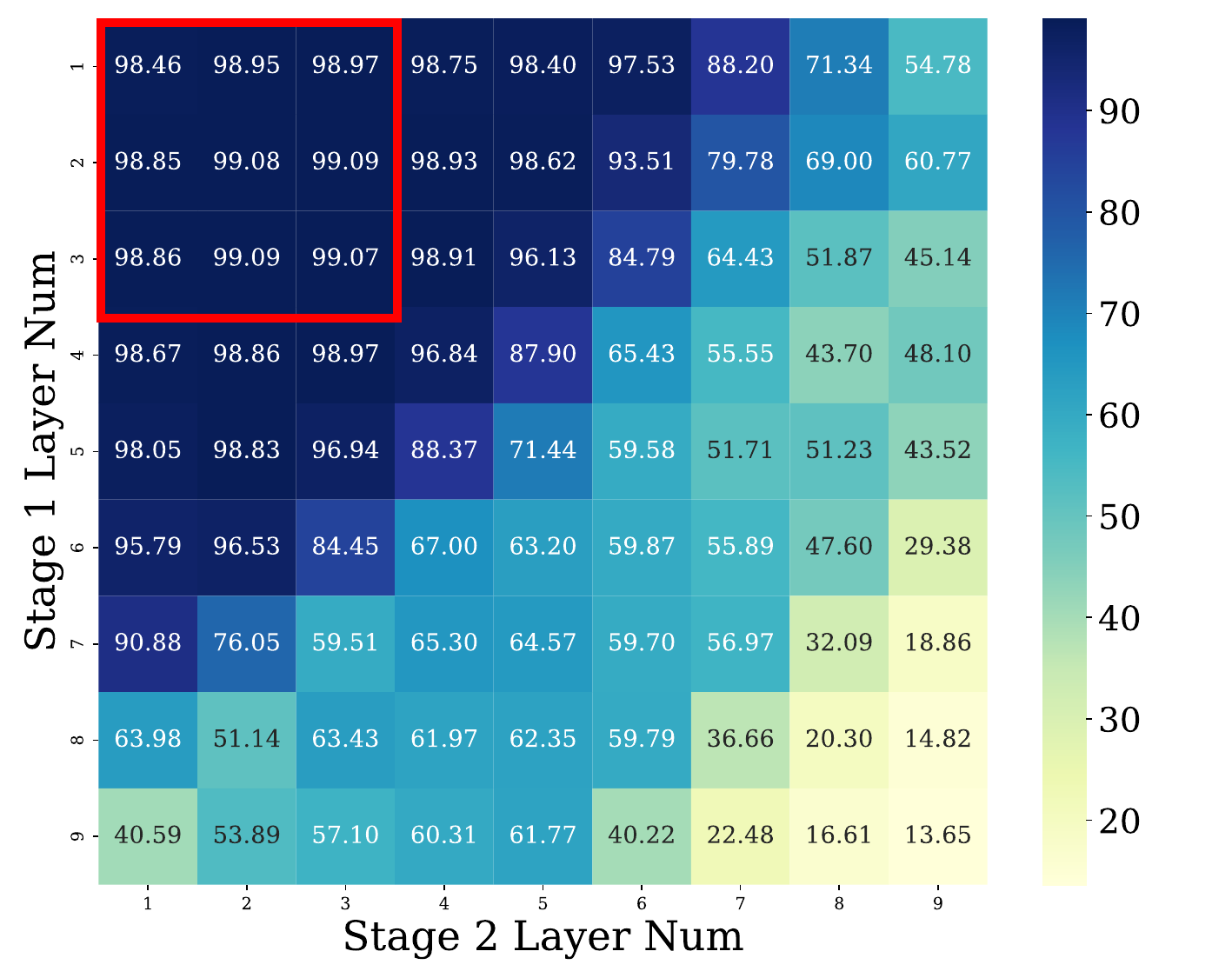}
    \caption{Experiments on Depth Morphing. Configurations enclosed in the \textcolor{red}{red} rectangle represent those encountered during training. All other configurations displayed were not included in the training phase.}
    \label{fig:Depth}
\end{figure}

\noindent\textbf{Results.} The performance are reported in Figure~\ref{fig:Depth} as a heatmap. As expected, the models maintain a comparable performance, approximately $99\%$ accuracy, across different depth configuration $L1, L2 \in \{1,2,3\}$. Interestingly, we observe that \texttt{NeuMeta} is capable of smoothly extrapolating to unseen architectural configurations. However, performance largely dropped when the architecture significantly deviates from the trained setup. For instance, with a ResNet configuration of $L1=L2=9$, the accuracy drops to just 13\%. All together, it shows that our \texttt{NeuMeta} is not confined to width morphing alone, showing further exploration direction.

\section{Weight Permutation Strategy}
\label{sec:permute}
In the main paper, we have reformulated the within-network smoothness problem as a weight permutation problem, solved through a multi-objective Shortest Hamilton Path~(mSHP) Problem. This section compares our method with the existing layer-wise Traveling Salesman Problem (TSP) permutation strategies.
\begin{table}[th]
    \centering
    \begin{tabular}{l|c|c}
    \toprule
        Method &  Accuracy &  TV\\
        \midrule
        Original (No Permute) & 91.62 & 58141.80\\
        \midrule
        Layer-TSP Greedy & 91.67 & 50271.82\\
        Layer-TSP 2-Opt & 91.31 & 49132.78\\
        \rowcolor{cyan!10} mSHP~(Ours) & \textbf{91.87} & \textbf{45576.02}\\
        \bottomrule
    \end{tabular}
    \caption{Ablation results with different weight permutation strategies.}
    \label{tab:permutation_cifar}
\end{table}

\begin{figure}[!h]
    \centering
\includegraphics[width=0.4\linewidth]{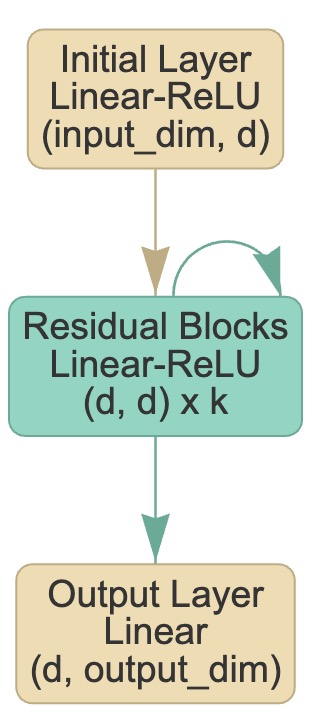}
    \caption{Prototype of network architecture.}
    \label{fig:arch}
\end{figure}
\noindent\textbf{Baselines and Measurement.} For comparison, we consider the original, unpermuted weights and three different permutation strategies. The first is the \emph{Random} permutation, which shuffles the weights based on a randomly generated order. We also compare our results with two layer-wise TSP strategies: the \emph{2-Opt} algorithm~\cite{Solodskikh_2023_CVPR}, and a \emph{Greedy} solver approach~\cite{DBLP:conf/iclr/AshkenaziRVLRMT23}.
\begin{figure*}[htbp]
    \centering
    \includegraphics[width=\linewidth]{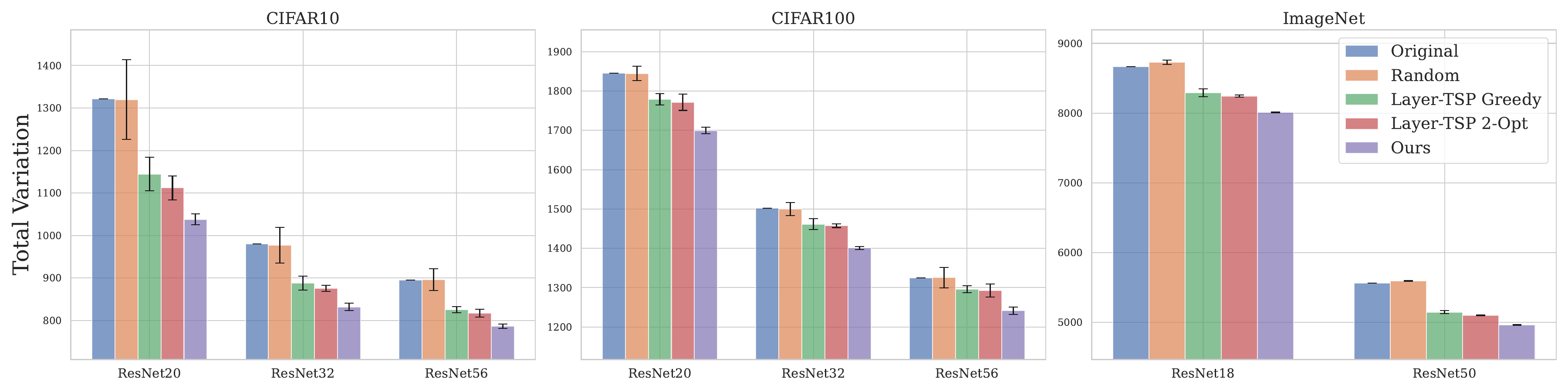}
    \caption{The average total variation for model weights after different permutation strategies. We show the mean$\pm$ over 5 runs.}
    \label{fig:smooth-tv}
\end{figure*}
Starting from the same trained network, we apply each permutation algorithm and measure the resulting total variation, averaged across all parameters. We repeat these experiments five times for each strategy and report the mean and standard deviation (mean $\pm$ std). Besides, we train the INR model on weights modified by these permutations using the CIFAR10 dataset, to assess the impact of weight permutation on final model performance.

\noindent\textbf{Results.} The results are presented in Figure~\ref{fig:smooth-tv}. We evaluated our mSHP solution across eight different network architectures on three datasets. The results demonstrate that our mSHP approach consistently achieves the best performance in terms of TV score after permutation. These results are highlighted in \textcolor{purple}{Purple} in the figure. 

Furthermore, we investigated the performance of permuted weights on the CIFAR10 dataset, as detailed in Table~\ref{tab:permutation_cifar}. Our finding indicates that compared to existing permutation strategies, our approach yields a slight improvement in performance.

\section{Additional Ablation Studies}
\label{sec:ablation}

\subsection{Model Architecture}

In the paper, we utilize a residual MLP with ReLU activation, as depicted in Figure~\ref{fig:arch}. The input dimension, represented as \(\textsc{input\_dim}\), is computed as \((\textsc{num\_freq} \times 2 + 1) \times 6\). This formulation indicates that the input is expanded by Fourier features. In this section, our discussion mainly revolves around three critical parameters of the network: the number of layers (\(k\)), the number of frequencies, the hidden channel number (\(d\)), and the usage of residual connection.

\noindent\textbf{Experiment Setup.}
In our experimental analysis, we explored the impact of varying model architectures on accuracy. The primary variables altered in these experiments were the number of layers, the number of hidden units, and the number of frequencies within the model. Specifically, we tested models with 5, 8, and 12 layers, each with either 256 or 512 hidden units, and frequency counts of 16 and 32. Additionally, we conducted comparisons between models with and without residual connections to understand their impact on performance.

\noindent\textbf{Results and Findings.}
As shown in Figure~\ref{tab:archi}, the results provided insights into the optimal model architecture. The best performance was observed in models with 8 layers and 512 hidden units at 32 frequencies, achieving a peak accuracy of 91.87\%. Models with 5 layers and 256 hidden units also performed well, particularly at 16 frequencies, reaching an accuracy of 91.76\%. However, models with 12 layers and 256 hidden units experienced a decline in accuracy, regardless of frequency count. This indicates that a configuration of 8 layers, 512 hidden units, and 32 frequencies is most conducive to high accuracy in our model setup.

The results comparing performances with and without residual connections are presented in Table~\ref{tab:residual}. It is observed that incorporating residual connections enhances performance by an increase of 1.69\% in accuracy.

\begin{table}[]
    \centering
    \begin{tabular}{l|l|l|c}
    \toprule
        Num Layer & Num Hidden & Num Freq & Accuracy \\
        \midrule
        5 & 256 & 16 & 91.76 \\
        5 & 256 & 32 & 91.23 \\
        8 & 256 & 16 & 91.68  \\
        8 & 256 & 32 & 91.27 \\
        8 & 512 & 16 & 91.84 \\
        \rowcolor{cyan!15}  8 & 512 & 32 & \textbf{91.87} \\
        12 & 256 & 16 & 91.27  \\
        12 & 256 & 32 &  91.03 \\
         \bottomrule
    \end{tabular}
    \caption{Ablation study on model architectures hyper-parameter.}
    \label{tab:archi}
\end{table}

\begin{table}[]
    \centering
    \begin{tabular}{l|c}
    \toprule
        Method &  Accuracy\\
        \midrule
        W/o residual & 90.18\\
       \rowcolor{cyan!15} W residual & \textbf{91.87}\\
        \bottomrule
    \end{tabular}
    \caption{Ablation study on residual connection.}
    \label{tab:residual}
\end{table}

\subsection{Block-Based INR}

In our efforts to enhance the capacity of the implicit function, we have implemented a block-based INR. This approach involves designating distinct MLP networks for predicting the different parameters (each \texttt{weight} or \texttt{bias}). The effectiveness of this method is assessed by comparing it against a conventional approach, where a singular INR is utilized to predict the weights for all layers.

\begin{table}[tbh]
    \centering
    \begin{tabular}{l|c}
    \toprule
        Method & Accuracy \\
        \midrule
         W/o block-based INR & 91.74\\
        \rowcolor{cyan!15} W block-based INR & \textbf{91.87} \\
         \bottomrule
    \end{tabular}
    \caption{Ablation study on block-based INR.}
    \label{tab:block}
\end{table}

\noindent\textbf{Results.} The comparison is illustrated in Table~\ref{tab:block}. Our finding indicates that employing the block-based INR strategy leads to enhancement in performance. Specifically, there is an observed improvement of 0.13\% in accuracy on the CIFAR10 dataset when using the block-based INR, highlighting its efficacy.

\subsection{Exponential Moving Average (EMA)}

In our study, we incorporate the Exponential Moving Average~(EMA) technique during the training phase of the INR. This approach updates the weights of the INR using  $\theta^t\leftarrow \gamma \theta^{t-1} + (1-\gamma) \theta^t$. We would like to evaluate the efficiency of the EMA technique and determining the optimal value for the $\gamma$ parameter.

\begin{table}[thb]
    \centering
    \begin{tabular}{l|c}
    \toprule
        $\gamma$ & Accuracy (\%) \\
        \midrule
        0 & 91.32\\
        0.99 & 91.42\\
        \rowcolor{cyan!15} 0.995 & \textbf{91.87} \\
        0.999 & 91.42\\
        \bottomrule
    \end{tabular}
    \caption{Ablation study evaluating different $\gamma$ values in EMA.}
    \label{tab:ema}
\end{table}

\noindent\textbf{Results.} The results of our ablation study, as shown in Table~\ref{tab:ema}, reveal the impact of varying $\gamma$ values on the accuracy of the model. It is evident that the optimal value of $\gamma=0.995$ significantly enhances the model's performance.

\section{Distinction From NeRN}
\label{sec:comparenern}
While there are work that apply INR to fit neuron weight like Neural Representation for Neural Networks (NeRN)~\cite{DBLP:conf/iclr/AshkenaziRVLRMT23}, our method, \texttt{NeuMeta}, stands distinct in several key aspects:
\begin{itemize}
    \item \textbf{Memorize \textit{vs}. Generalize.} NeRN is tailored to fit a \emph{single} network, and lacks the ability to extrapolate to unseen architectures post-training. In contrast, \texttt{NeuMeta} is designed to adapt to the \emph{entire manifold}. Once trained, it can generate weight for any network configuration on this manifold, without retraining.
    \item \textbf{Discrete \textit{vs}. Continuous.} Unlike NeRN, which fits a discrete network with coordinate-wise inputs, \texttt{NeuMeta} embraces a continuous manifold approach. This allows weight values  to be represent by an average over a small neighborhood, enhancing generalization to unseen architectural weights beyond the training scope.
    \item \textbf{Purpose.} NeRN focuses on model compression, storing the parameters of a full model within a smaller MLP to reduce the parameter count. \texttt{NeuMeta}, however, aims for resizeability and flexibility, enabling on-the-fly sampling of different network weights.
\end{itemize}

Given these factors, our \texttt{NeuMeta} represents a significant advancement beyond NeRN, moving past the confines of fitting weights for a single, discrete network. 

\section{Segmentation Visualization}
\label{sec:vis}
\begin{figure*}
    \centering
    \includegraphics[width=\linewidth]{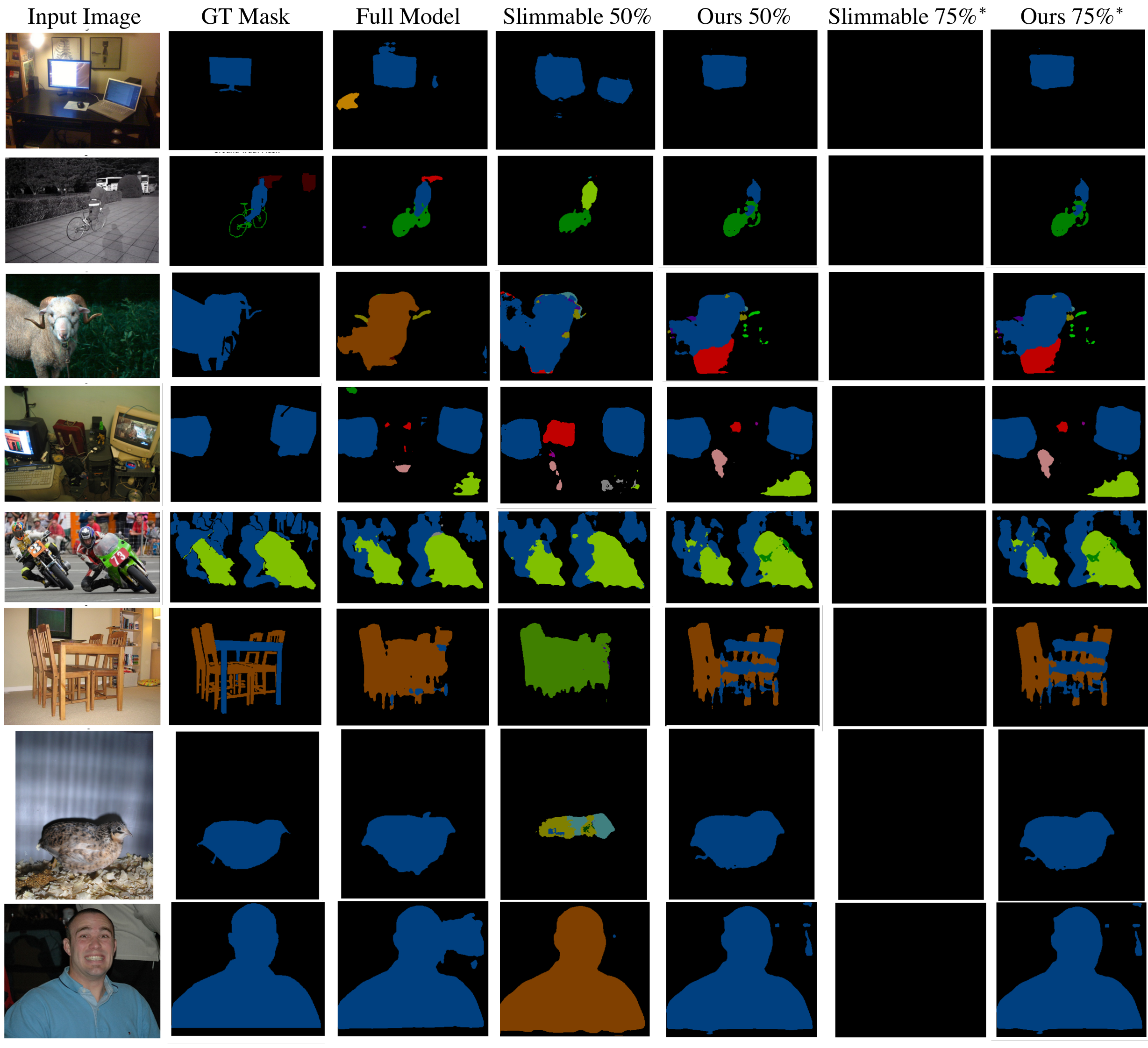}
    \caption{Semantic segmentation visualization on VOC2012 dataset. $*$ denotes that the model has not been trained with a $75\%$ compression rate.}
    \label{fig:visualiz}
\end{figure*}

Building upon our performance analysis of semantic segmentation in the main paper, we offer an in-depth qualitative comparison in Figure~\ref{fig:visualiz}. This includes a comparison with the full-sized model (shown in column 3) and models trained using the Slimmable Neural Network approach (columns 4 and 6).

We make two major observations
\begin{itemize}
    \item \textbf{Enhancement Over Slimmable NN.} Notably, our \texttt{NeuMeta} consistently outperforms the Slimmable NN baseline. In the $50\%$ compression scenario, \texttt{NeuMeta} achieves more accurate mask predictions. Even in the challenging $75\%$ compression case, where Slimmable NN struggles significantly, \texttt{NeuMeta} manages to produce reasonable segmentation outputs.
    \item  \textbf{Comparison with Full-Sized Model}: Interestingly, \texttt{NeuMeta} not only competes with but also surpasses the full-sized model in several instances, such as in rows 3, 6, and 8. 
    These observations might indicate the potential efficacy of \texttt{NeuMeta}, especially in terms of how its smoothed weight could improve the model's generalization capabilities across various challenging contexts.
\end{itemize}

These findings underscore \texttt{NeuMeta}'s potential in providing more accurate semantic segmentation, particularly under large compress rate, demonstrating its value in enhancing model performance and efficiency.

\section{Experimental Setup}

\subsection{Selection of Pretrained Models}
For ResNet models applied to CIFAR10 and CIFAR100 datasets, we utilize the models trained and available at~\footnote{\url{https://github.com/chenyaofo/pytorch-cifar-models}}. In the case of ResNet on the ImageNet dataset, we employ the official pretrained models provided by PyTorch. For tasks involving MNIST, image generation, and VOC segmentation, we have trained the models ourselves and subsequently integrated them into our Neural Metamorphosis pipeline.

\subsection{Task-specific Setup and Selected Layers}

In our various experiments, we selectively applied \texttt{NeuMeta} to different layers of the network. This approach was adopted because we discovered that applying the INR to all layers simultaneously presents significant computational and convergence challenges. Below are the details of the layers selected for each task:

\begin{itemize}
    \item \textbf{Classification on MNIST}: For MNIST experiments with LeNet, we apply fitting to all layers, with 7 parameters to be fit.
    \item \textbf{Classification on CIFAR10 and CIFAR100}: In these cases, we utilize the Implicit Neural Representation (INR) to fit only the last residual block of ResNet20, specifically \texttt{layer.3.2}, with 4 parameters to be fit.
    \item \textbf{Classification on Imagenet with ResNet18}: Here, the INR is used to fit the last residual block of stage 3, denoted as \texttt{layer.3.1}, with 4 parameters to be fit.
    \item \textbf{Segmentation on VOC2012}: We employed U-Net alongside ResNet18 as our backbone for segmentation on VOC2012. Here, INR was used to fit the last residual block of stage 3, \texttt{layer.3.1}, with 4 parameters to be fit.
    \item \textbf{VAE on MNIST}: We utilized a fully connected network for image generation. Each image was flattened into a 1D vector, fitted in the model, compressed to mean and variance, and reconstructed. INR was used to fit the last two layers of this VAE, specifically \texttt{fc5} and \texttt{fc6}, with 4 parameters to be fit.
    \item \textbf{VAE on CelebA}: For generating facial images from CelebA at a resolution of $64\times 64$, we used a convolutional VAE. INR was applied to the last three convolution layers, namely \texttt{decoder.3}, \texttt{decoder.6}, and \texttt{decoder.9}, with 6 parameters to be fit in total.
\end{itemize}

\section{Limitations and Future Directions}
While our pipeline is designed to be straightforward and broadly applicable, it encounters several intrinsic limitations in terms of scalability and practicality:

\begin{itemize}
    \item \textbf{Scale to All Layers.} Our initial attempt to use a single Implicit Neural Representation (INR) to fit all layers' parameters often resulted in slow training and non-convergence. This issue is attributed to the large size of network parameters. For instance, ResNet18 contains 11.68 million parameters, making the manifold extremely vast when sampling hundreds of networks. Fitting a single MLP to this extensive number of input-output pairs proves challenging. We offered a block-based solution as a workaround, but the underlying issue remains.
    \item \textbf{Performance Limitation with Full-Sized Model.} Our experiments indicate a marginal decrease in performance when using \texttt{NeuMeta} on the full-sized model. This limitation primarily arises from the inadequate representational capability of the INR in substituting neural weights. The inherent smoothness bias of the INR poses a challenge, making it difficult to precisely reconstruct the full model's architecture.

    \item \textbf{Dealing with BatchNorm Layers.} To accommodate BatchNorm (BN) layers in our INR, we converted them into linear operations. However, this approach complicates the training of deeper networks, especially considering the critical role of BN in training such networks. A more sophisticated method is required to represent BN layers effectively with an implicit function.

    \item \textbf{Easy Extension of INR.} We employ a basic version of Implicit Neural Representation (INR) in our study, using a simple MLP, Fourier position encoding, and ReLU activation. The field of INR, particularly its application to various signals, is rapidly evolving. Our approach highlights the scope for exploring and implementing more advanced INR architectures in future research.
\end{itemize}

These limitations highlight areas for future refinement and underscore the need for innovative approaches to address scalability and the representation of complex network components.

\section{Source Code}
For your reference, the full source code is provided in the attached zipped file named \textsc{Neural-Metamorphosis-code.zip}. The code is fully documented. Please refer to the code for detailed information and implementation details.

{
    \small
    \bibliographystyle{ieeetr}
    \bibliography{main}
}

%% file: sec/0_abstract.tex
\begin{abstract}
This paper introduces a new learning paradigm 
termed \textbf{Neural Metamorphosis}~(\textbf{NeuMeta}), 
which aims to 
build self-morphable neural networks. 
Contrary to crafting separate models 
for different architectures or sizes, 
\texttt{NeuMeta} directly learns the continuous \textit{weight manifold}
of neural networks. 
Once trained, we can sample weights 
for any-sized network directly from the manifold, even for previously unseen configurations,
without retraining. 
To achieve this ambitious goal,
\texttt{NeuMeta} trains neural implicit functions as hypernetworks. They accept coordinates within the \textit{model space} as input, 
and generate corresponding weight values on the manifold.
In other words, 
the implicit function is learned 
in a way, that the 
predicted weights is well-performed across various models sizes. 
In training those models, 
we notice that, the final performance 
closely relates on smoothness of the learned manifold. In pursuit of enhancing this smoothness, we employ two strategies. First, we permute weight matrices to achieve intra-model smoothness, 
 by solving the Shortest Hamiltonian Path problem. Besides, we add a noise on the input coordinates when training the implicit function, ensuring models with various sizes shows consistent outputs. As such, \texttt{NeuMeta} shows promising results in synthesizing parameters for various network configurations. Our extensive tests in image classification, semantic segmentation, and image generation reveal that \texttt{NeuMeta} sustains full-size performance even at a 75\% compression rate.

\keywords{Weight Manifold \and Morphable Neural Network \and Implicit Neural Representation}
\end{abstract}

%% file: sec/1_intro.tex
\section{Introduction}
\label{sec:intro}

The world of neural networks is mostly 
dominated by the \emph{rigid} principle: Once trained, they function as static monoliths with immutable structures and parameters. Despite the growing intricacy and sophistication of these architectures over the decades, this foundational approach has remained largely unchanged.

This inherent rigidity presents challenges, 
especially when deployed 
in new scenarios 
unforeseen during the network's initial design. 
Each unique scenario calls for a new model of distinct configuration, involving repeated design, training, and storage processes. Such an approach is not only resource-intensive, but also limits the model's prompt adaptability in rapidly changing environments.



In our study, we embark on an ambitious quest to design neural networks that can, once trained, be continuously morphed for various hardware configurations. Particularly, our goal is to move beyond the confines of fixed and pre-trained architectures,
and create networks that readily generalize to 
unforeseen sizes and configurations during the training phase.

Indeed, this problem has been considered in slightly different setup, , employing strategies like flexible models ~\cite{cai2020once,DBLP:conf/iclr/YuYXYH19,devvrit2023matformer} and network pruning techniques~\cite{DBLP:conf/iclr/0022KDSG17,DBLP:conf/iclr/FrankleC19,fang2023depgraph}.
The former ones 
are designed to self-adapt to various subnetwork configurations, 
whereas the latter ones aim to eliminate redundant connections, 
achieving models that are streamlined yet robust. 
Nevertheless, these solutions have their own challenges: 
flexible models are confined to the their training configurations, 
and pruning methods
compromise performance 
and often require further retraining. 
Most importantly,
they are still building 
numerous rigid models, without the ability to be continuously morphed.

\setlength{\intextsep}{0pt}%
\setlength{\columnsep}{0pt}%
\setlength{\floatsep}{0pt}
\setlength{\textfloatsep}{0pt}
\begin{wrapfigure}{r}{0.53\textwidth}
  \begin{center}
    \includegraphics[width=\linewidth]{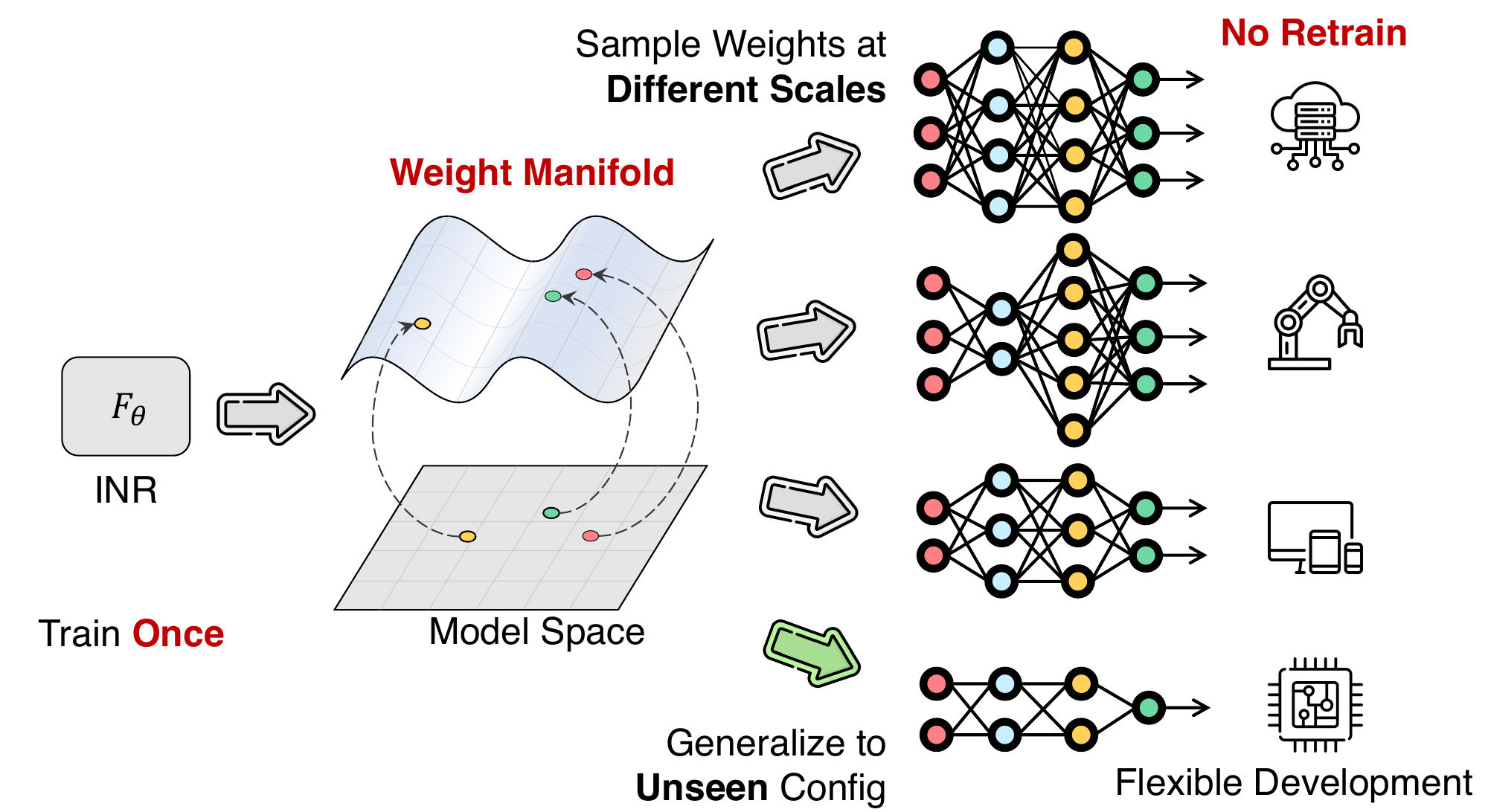}
  \end{center}
  \caption{Pipeline of \textbf{Neural Metamorphosis}. }
    \label{fig:pipeline}
\end{wrapfigure}

To this end, we present a new learning paradigm, termed \textbf{Neural Metamorphosis}~(\texttt{NeuMeta}). That is, instead of interpreting neural networks not as discrete entities, we see them as points sampled from a continuous and high-dimensional \emph{weight manifold}. This shift allows us to learn the manifold as a whole, rather than handling isolated points. As such, \texttt{NeuMeta}, as its name implies, can smoothly morphs one network to another, with similar functionality but different architecture, such as width and depth. Once done, we can generate the weights for arbitrary-sized models, by sampling directly from this manifold.





At the heart of our paradigm is the use of Implicit Neural Representation (INR) as hypernetworks to fit the manifold.
Intuitively, the INR acts as an indexing function of weights: upon receiving the network configuration and weight coordinates as inputs, it produce the corresponding weight values. In the training phase, the INR is assigned two goals: it approximates the weights of the pre-trained network, while simultaneously minimizing the task loss across a variety of randomly sampled network configurations. During the testing phase, the INR receives the weight coordinates for the desired network configuration, outputting values to parameterize the target network. This implementation is, by nature, different from existing methods that builds continuous neural networks that rely on integral operations and explicit continuous weight function~\cite{Solodskikh_2023_CVPR, le2007continuous}.

Nonetheless, our ambitious effort comes with great challenges, causing the simplistic solutions to fail. A primary difficulty arises from the inherently non-smooth properties of network's weights, which hinders the fitting of the INR. 

To overcome this, we put forth two strategic solutions. The first involves the permutation of weight matrices to enhance the \textbf{intra-network smoothness}. Recognizing the flaws of prior attempts, we formulate it as a \emph{multi-objective Shortest Hamiltonian Path problem}~(mSHP). By solving this problem within individual network cliques, we enhance the smoothness of the network's weights.

The second strategy, aimed at \textbf{cross-network smoothness}, involves introducing random noise into the input. During the INR training, we maintain output consistency of the main model, irrespective of this input noise. During testing, the expected output value around this sampled coordinates is employed as the predicted weight value. This strategy moves away from a rigid grid, allowing for greater flexibility in generating new networks. Together, these strategies simplify the process of modeling the weight manifold, thereby enhancing the robustness of our approach.


We evaluated \texttt{NeuMeta} on various tasks including image classification, semantic segmentation, and generation. Our findings reveal that \texttt{NeuMeta} not only matches the performance of original model but also excels across different model sizes. Impressively, it maintains full-size model performance even with a 75\% compression rate. Remarkably, \texttt{NeuMeta} extrapolates unseen weights. In other words, it can  generates parameters for network sizes outside its training range, accommodating both larger and smaller models.

This paper's contributions are summarized as follows:
\begin{itemize}
    \item We introduce Neural Metamorphosis, a new learning paradim that leverage INRs to learn neural networks' continuous weight manifold. Once trained, this INR can generate weights for various networks without retraining.
    \item We introduce dual strategies to improve both \emph{intra-network} and \emph{cross-network} smoothness of the weight manifold. This smoothness is key to generate high-performing networks.
    \item The proposed method undergoes thorough evaluations across multiple domains, including image classification, segmentation and generation, underscoring their versatility and robustness in varied computational setup.
\end{itemize}

%% file: sec/1_1_related.tex
\section{Related Work}

\textbf{Efficient Deep Neural Networks.} In recourse-limited applications, the efficiency of neural networks becomes a critical concern. Researcher have explored structure pruning methods~\cite{DBLP:journals/corr/HanMD15,ma2023llm} that trim non-essential neurons to reduce computation. Another development is the flexible neural networks~\cite{DBLP:conf/iclr/YuYXYH19,yu2019universally,cai2020once,grimaldi2022dynamic,chavan2022vision,hou2020dynabert}, which offer modifiable architectures. These networks are trained on various subnetwork setups, allowing for dynamic resizing. In our context, we present a new type of flexible model that directly learns the continuous weight manifold. This allows it to generalize to configurations that haven't been trained on, an infeasible quest for existing approaches.
\\\noindent\textbf{Continuous Deep Learning.} Beyond traditional neural networks that discretize weights and outputs, continuous deep learning models represent these elements as continuous functions~\cite{chen2018neural,Solodskikh_2023_CVPR}. The concept extends to neural networks with infinite hidden width, modeled as Gaussian processes~\cite{neal2012bayesian}. Further endeavors replace matrix multiplication with integral operations~\cite{DBLP:journals/jmlr/RouxB07,Solodskikh_2023_CVPR}. Neural Ordinary Differential Equations (Neural ODEs)~\cite{chen2018neural} build models by defining dynamical system with differential equations, fostering a continuous transformation of data. Our method diverges from those above, using an INR to create a continuous weight manifold. This allows for continous weight sampling and introduces a new type of continuous network.

\setlength{\floatsep}{2pt}
\setlength{\textfloatsep}{2pt}
\setlength{\intextsep}{2pt}
\begin{table}[t!]
    \centering
\renewcommand{\arraystretch}{1.}
\tiny
    \begin{tabular}{lccccc} 
        \toprule 
        \textbf{Method} & \textbf{Continuous} & \textbf{HyperNet} & \textbf{Resizable} & \textbf{Checkpoint-Free} & \textbf{Generalize to Unseen} \\ 
        \midrule 
        Structure Prune~\cite{molchanov2019importance}     & \xmark & \xmark & \cmark$^{*}$ & \cmark & \cmark  \\ %
        Network Transform~\cite{wei2016network,DBLP:journals/corr/ChenGS15,yang2022deep}  & \xmark & \xmark & \cmark & \cmark & \cmark  \\ 
        Flexiable NN~\cite{cai2020once,DBLP:conf/iclr/YuYXYH19} & \xmark & \xmark & \cmark & \cmark & \xmark\\ 
        Continuous NN~\cite{Solodskikh_2023_CVPR,DBLP:journals/jmlr/RouxB07}       & \cmark & \xmark & \cmark & \cmark & \xmark \\ 
        Weight Generator~\cite{knyazev2021parameter}   & \xmark & \cmark & \cmark& \xmark & \cmark \\ 
        \midrule 
        \texttt{NeuMeta}~(Ours) & \cmark & \cmark & \cmark & \cmark & \cmark\\ 
        \bottomrule 
    \end{tabular} 
    \caption{Comparing methods for building neural networks but can be resized. Structural pruning ($*$) only reduces network size. Network Transform manipulates weights to construct a functionally identical version of the source network. Flexible Models are training-dependent and fail with unseen networks. Existing Continuous NNs are only valid for specific operators. Weight generators need extensive training checkpoints. Our \texttt{NeuMeta} uniquely learns a continuous weight manifold with INR hypernet. As such we can generalize to any neural operator and unseen configurations beyond training.} 
    \label{tab:compare} 
\end{table}
\noindent\textbf{Knowledge Transfer.} Pre-trained neural networks have become a cornerstone for advancing deep learning, enabling rapid progress in downstream tasks due to their transferable learned features~\cite{wei2016network,DBLP:journals/corr/ChenGS15}. Techniques like network transform~\cite{yang2022deep,wei2016network,DBLP:journals/corr/ChenGS15} and knowledge distillation~\cite{hinton2015distilling,yang2022factorizing,yuan2020revisiting} adapt these features to fit new architectures and more compact networks. Our approach also transfer knowledge, but instead of doing a one-to-one transfer, we derive a continuous manifold of weights from a trained neural network. This enables a one-to-many knowledge transfer, which can create multiple networks of various sizes.
\\\noindent\textbf{HyperNetworks.} HyperNetworks optimize neural architectures by generating weights via auxiliary networks~\cite{45803}. To accommodate weights for new models or architectures, they are trained on a vast range of checkpoints, learning a weight distribution~\cite{knyazev2021parameter,schurholt2022hyper,peebles2022learning}, facilitating multitask learning~\cite{navon2021learning,raychaudhuri2022controllable}, continual learning~\cite{Oswald2020Continual}, fewshot learning~\cite{sendera2023hypershot} and process implicit function~\cite{deluigi2023inr2vec}. 

Unlike typical hypernetworks producing fix-sized weights, our method uses an INR as hypernetwork that learns to predict \textit{variable-sized} weights, offering dynamic, on-demand weight adaption. Similarly, \cite{DBLP:conf/iclr/AshkenaziRVLRMT23} uses INR as hypernet, but their approach is confined to fixed-size weight prediction. \cite{raychaudhuri2022controllable} predicts connections between layers without accounting for various weight sizes.

We provide a comparative analysis of the aforementioned methods in Table~\ref{tab:compare}.

%% file: sec/4_method.tex
\section{Implicit Representation on Weight Manifold}

In this section, we introduce the problem setup of \texttt{NeuMeta} and present our solution. In short, our idea is to create a neural implicit function to predict weight for many different neural networks. We achieve this goal by posing the principle of smoothness in this implicit function.


\subsection{Problem Definition}

Let's imagine the world of neural networks as a big space called \(\mathcal{F}\). In this space, every neural network model \(f_\mathbf{i} \in \mathcal{F}\) is associated with a set of weight \(\mathbf{W}_\mathbf{i} = \{w_{(\mathbf{i},\mathbf{j})}\}\). Such a model $f_\mathbf{i}$ is uniquely identified by its configuration \(\mathbf{i} \in \mathcal{I}\), such as width (channel number) and depth (layer number). Furthermore, each weight element within the network is indexed by \(\mathbf{j} \in \mathcal{J}\), indicating its specific location, including aspects like the layer index and channel index. The combination of configurations and indices, \(\mathcal{I} \times \mathcal{J}\) forms the \emph{model space}, uniquely indexing each weight. We say the all weights values that makes up a good model on a dataset $D$ lies on a \emph{weight manifold} $\mathcal{W}$. We also assume we have access to a pretrained model $f(\cdot;\mathbf{W}_{\text{original}})$. Our goal is to learn the weight manifold $\mathcal{W}$.

\noindent\textbf{Definition 1. (Neural Metamorphosis)} \textit{ 
 Given a labeled dataset $D$ and a pretrained model $f(\cdot;\mathbf{W}_{\text{original}})$, we aim to develop a function $F: \mathcal{I} \times \mathcal{J} \rightarrow \mathcal{W}$ that maps any points in the model space to its optimal weight manifold. This is achieved by minimizing the expected loss across full $\mathcal{I} \times \mathcal{J}$.}
{\small\begin{equation}
    \min_{F} \mathbb{E}_{\forall (\mathbf{i},\mathbf{j}) \in \mathcal{I} \times \mathcal{J}}\big[\mathcal{L}_{\text{task}}(f_\mathbf{i}(\mathbf{W}^*_{\mathbf{i}});D)\big],\quad
    s.t. \mathbf{W}^*_{\mathbf{i}} = \{w^*_{(\mathbf{i},\mathbf{j})}\}, 
    w^*_{(\mathbf{i},\mathbf{j})}=F(\mathbf{i},\mathbf{j}), \label{eq:opt}
\end{equation}}

where $\mathcal{L}_{\text{task}}$ denotes the task-specific loss function. In other words, $F$ give us the best set of weights for any model setup in \(\mathcal{I} \times \mathcal{J}\), rather than fitting a single or a set of neural networks~\cite{cai2020once,DBLP:conf/iclr/YuYXYH19}. 
In context neural network, our $F$, which inputs coordinates and outputs values, is known as implicit neural representation~(INR)~\cite{sitzmann2020implicit}. Consequently, we choose INR as our $F$, offering a scalable and continuous method learn this mapping.




\noindent\textbf{Connecting to Continuous NN.} \texttt{NeuMeta} can be viewed as a method to build Continuous NNs, by representing weight values as samples from a continuous weight manifold. Here, we would like to see how it differs from existing methods. For example, in continuous-width NNs~\cite{DBLP:journals/jmlr/RouxB07,Solodskikh_2023_CVPR}, linear operations are typically defined by the Riemann integral over inputs and weights:
{\small \begin{equation}
    f(x)= \Vec{\mathbf{x}}\cdot\Vec{\mathbf{w}} = \sum_{j} (\Delta_j x_j W(j) ) \approx \int_{0}^1 x(j) W(j) \delta j, \label{eq:integral}
\end{equation}}
where $\Vec{\mathbf{x}}$ and $\Vec{\mathbf{w}}$ represent discrete input and weight vectors. $j$ is the continuous-valued channel index. $W(j)$ is a continuous weight function, and $\Delta_j$ is the width of the sub-interval between sampled points for integral. 

Our method offers three key advantages over this traditional approach:
 \begin{itemize}
    \item \underline{Integral-Free.} \texttt{NeuMeta} requires no integral.
     \item \underline{Learned Continuous Sampling.} Our method jointly learns the continuous weight function and the sampling interval $w_j=\Delta_j W(j)$, rather that learning $W(j)$ along. This enables us to generate continuous-width NN on-fly, a feat unachievable with discrete learned sampling~\cite{Solodskikh_2023_CVPR}.
     \item \underline{INR Parameterization.} INR offers a generalized form to model the continuous function\footnote{Prior designs using kernel~\cite{Solodskikh_2023_CVPR} or piece-wise functions~\cite{DBLP:journals/jmlr/RouxB07} can be considered special cases of INR, as detailed in our supplementary material.}.
 \end{itemize}

\begin{figure}[!thb]
    \centering
    \includegraphics[width=0.8\linewidth]{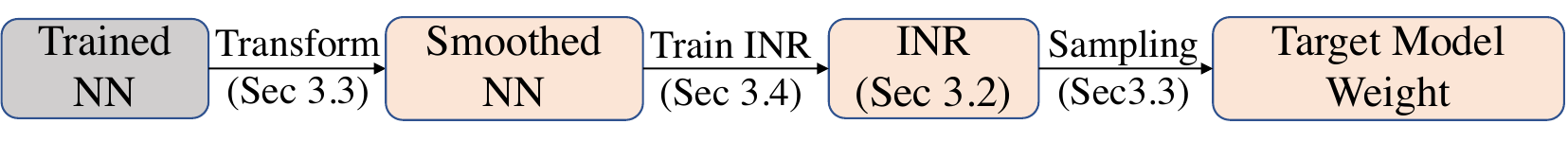}
    \caption{Diagram of \texttt{NeuMeta} and our content organization.}
    \label{fig:pipeline}
\end{figure}
\noindent\textbf{Challenge and Solution.} Our effort, while ambitious, presents distinct challenges. First, an INR design for neural network weight is largely unexplored. Second, it is essential to train on limited samples from the weight manifold and then generalize to unseen ones. Our solution, as depicted in Figure~\ref{fig:pipeline}, includes an INR-based architecture in Section~\ref{sec:arch} and a strategy for learning on a smooth weight manifold in Sec~\ref{sec:smooth}. The training process is discussed in Sec~\ref{sec:train} .

\subsection{Network Architecture}
\label{sec:arch}
At the core of \texttt{NeuMeta}, we employ an INR model, $F(\cdot;\theta): \mathbb{R}^{k} \to \mathbb{R}^{d}$, to parameterize the weight manifold. This function, based on a multi-layer perceptron (MLP), transforms model space into weight values. In our implementation, we set the parameter $k=6$. For convolutional networks, the dimension $d=K \times K$, the maximum kernel size, whereas for non-convolutional setups, $d=1$.

Considering a generalized network with $L$ layers, each layer with an input-output channel size of $(C_{\text{in}}, C_{\text{out}})$. Each weight element, $w_{(\mathbf{i},\mathbf{j})}$, is associated with a unique index within the network. This index is represented as a coordinate pair $(\mathbf{i},\mathbf{j})$, with $\mathbf{i}=(L, C_{\text{in}}, C_{\text{out}})$ denoting the network structure and $\mathbf{j}=(l, c_{\text{in}}, c_{\text{out}})$ indicating the its specific layer, input, and output channel number. To ensure the same coordinate system is applicable to all $(\mathbf{i},\mathbf{j})$, these raw coordinates undergo normalization, typically rescaling them with a constant $N$
{\small\begin{align}
    \mathbf{v} &= \left[\frac{l}{L}, \frac{c_{\text{in}}}{C_{\text{in}}}, \frac{c_{\text{out}}}{C_{\text{out}}}, \frac{L}{N}, \frac{C_{\text{in}}}{N}, \frac{C_{\text{out}}}{N}\right],
\end{align}}
Similar to prior technique~\cite{mildenhall2021nerf}, the normalized coordinates undergo a transformation through sinusoidal position embedding, to extract its Fourier features.
{\small
\begin{align*}
    \gamma_{\text{PE}}(\mathbf{v}) &= \left[\sin(2^0\pi\mathbf{v}), \cos(2^0\pi\mathbf{v}), \ldots , \sin(2^{L-1}\pi\mathbf{v}), \cos(2^{L-1}\pi\mathbf{v})\right]
\end{align*}}
These encoded Fourier features, $\gamma_{\text{PE}}(\mathbf{v})$, serve as inputs to the MLP, yielding the weights:
{\small\begin{equation}
    w_{(\mathbf{i},\mathbf{j})} = F(\gamma_{\text{PE}}(\mathbf{v});\theta) = \frac{\text{MLP}(\gamma_{\text{PE}}(\mathbf{v});\theta)}{C_{\text{in}}}. \label{eq:1}
\end{equation}}
In equation (\ref{eq:1}), the output of the MLP is scaled by the number of input channels $C_{\text{in}}$, ensuring that the network's output maintains scale invariance relative to the size of the input channel~\cite{he2015delving,glorot2010understanding}. 

To handle lots of parameters with INR, we adopting a block-based approach~\cite{tancik2022block,reiser2021kilonerf}. Instead of a single large INR, weights are divided into a grid, with each segment controlled by a separate MLP network. The full architecture will be mentioned in the supplementary material.

In our framework, the weights for standard neural network operations are defined as follows:

\noindent\textbf{Linear Operation.}
For linear operations, we obtain the scalar weight as the element-wise average of $w_{(\mathbf{i},\mathbf{j})}$.

\noindent\textbf{Convolution Operation.}
For convolution layers, weights $w_{(\mathbf{i},\mathbf{j})}$ are reshaped into $k \times k$. If the kernel size $k$ is smaller than the $K$, only the central $k \times k$ elements are utilized.

\noindent\textbf{Batch Normalization Layer.}
 For batch normalization, we use a re-parameterization strategy~\cite{ding2021repvgg}, integrating BN weights into adjacent linear or convolution layers. This method integrates BN operations into a unified framework.

\subsection{Maintaining Manifold Smoothness}
A critical design within our paradigm is ensuring the weight manifold remains smooth.  We, in this section, discuss why this smoothness is crucial for the model's performance, and outline our strategy for achieving this local smoothness.


\label{sec:smooth}
\noindent\textbf{Intra-Model Smoothness.} Modern neural networks heavily rely on their ability to model smooth signals~\cite{rahaman2019spectral} to ensure convergence. Yet, empirical evidence suggests that the weight matrices are typically non-smooth. To enable our INR to reconstruct weights, we must find strategies that promote smoothness.

To address this challenge, previous studies have explored the concept of \emph{weight permutation}~\cite{Solodskikh_2023_CVPR,DBLP:conf/iclr/AshkenaziRVLRMT23}. It is often likened to the Traveling Salesman Problem (TSP)~\cite{lawler1986traveling}. 
However, such an approach, while seemingly straightforward, overlooks the crucial \emph{inter-dependencies} within and between weight matrices. 

\begin{figure}[t]
    \centering
    \includegraphics[width=\linewidth]{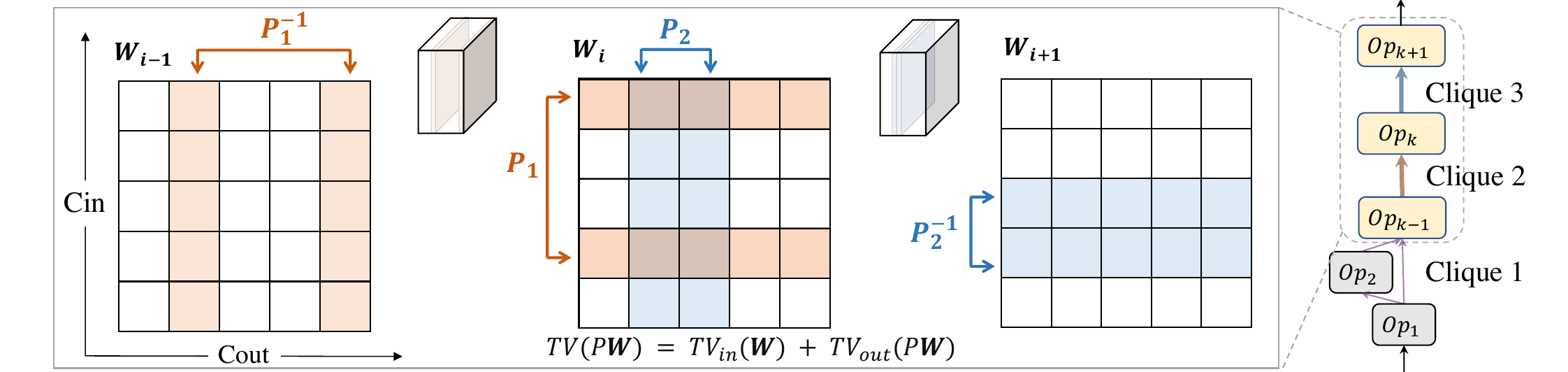}
    \caption{\textbf{Intra-model smoothness} via permutation equivalence. Our approach involves permuting weights to minimize total variance within each neural clique graph, thereby enhancing global smoothness.}
    \label{fig:permutation}
\end{figure}

Let's consider a weight matrix $\mathbf{W}\in \mathbb{R}^{ C_{\text{out}} \times C_{\text{in}}}$and measure its smoothness using \emph{total variation}, denoted as $TV(\mathbf{W})$. It is defined as the sum of variations along both channels: $TV(\mathbf{W}) = TV_{\text{in}}(\mathbf{W}) + TV_{\text{out}}(\mathbf{W})$. In fact, applying the TSP formulation presents 3 problems:

\begin{itemize}
\item \textbf{(P1) Loop VS Non-Loop}: Unlike TSP, which necessitates returning to the starting point, ensuring 2D weight matrix smoothness doesn't require looping back. Instead, it is better to be considered as a \emph{Shortest Hamiltonian Path}~(SHP)~\cite{DBLP:books/fm/GareyJ79} problem, allowing for an arbitrary starting channel.
\item \textbf{(P2) Breaking Smoothness for the Connecting Layer}: Unlike isolated weight matrices, neural networks consist of connected layers, creating complex inter-layer relationships. This is illustrated in Figure~\ref{fig:permutation}, where permutations in one layer necessitate corresponding reversals in adjacent layers to maintain the network's functional equivalence. For example, with an activation function $\sigma(\cdot)$ and a valid permutation pair $P$ and $P^{-1}$ (where $PP^{-1}=I$), the following equation holds:
{\small
\begin{align}
\mathbf{W}_i P\sigma(P^{-1}\mathbf{W}_{i-1}\mathbf{X}) = \mathbf{W}_i \sigma(\mathbf{W}_{i-1}\mathbf{X})
\end{align}}
As a result, $P^{-1}$ may affect the adjacent layers, with increased TV for $\mathbf{W}_i P$.
\item \textbf{(P3) Breaking Smoothness for the Other Dimension}: A permutation enhancing smoothness in output channel, might introduce non-smooth patterns in the input channel, thus reducing the overall smoothness.
\end{itemize}

Luckily, we find that the computation of the TV measurement renders (P3) infeasible, implying our focus should be directed towards (P1) and (P2).

\noindent\textbf{Proposition 1. (Axis Alignment)} \footnote{Proof in the supplementary material.} \textit{Let $\mathbf{W}$ be a given matrix and $P$ be a permutation. The application of a permutation in one dimension of $\mathbf{W}$ does not influence the total variation in the orthogonal dimension.}
{\small\begin{align}
TV(\mathbf{W}P) = TV_{\text{in}}(\mathbf{W}P) + TV_{\text{out}}(\mathbf{W})\\
TV(P\mathbf{W}) = TV_{\text{in}}(\mathbf{W}) + TV_{\text{out}}(P\mathbf{W})
\end{align}}

 Hence, to tackle global smoothness, we address challenges P1 and P2. We consider a neural network as a dependency graph $G=(V,E)$~\cite{fang2023depgraph}, where each node $v_i \in V$ represents an operation with weight $\mathbf{W}_i$ and each edge $e_{ij}\in E$ indicates inter-connectivity between $v_i$ and $v_j$. Each \emph{graph clique} $C=(V_C,E_C)\subset G$ is full-connected, representing a group of operation is connected. As a results, each $C$ corresponds to a unique permutation matrix $P$. Our objective is to determine all $P$ in a way that minimizes the total variation across the whole network. 
 
 Luckily, based on the Proposition 1, this complex optimization can be broken down into multiple independent optimizations, each on a clique. We define this as a multi-objective Shortest Hamiltonian Path (\emph{mSHP}) problem:
{\small \begin{align}
\underset{P}{\arg\min} \sum_{e_{ij} \in E_C} \Big(TV_{\text{out}}(P\mathbf{W}_i) + TV_{\text{in}}(\mathbf{W}_j P^{-1})\Big)
\end{align}}

To address each mSHP problem, we transform it into a TSP problem by adding a dummy node. This new node has edges with zero-distance to all others in the clique. We then solve TSP using a 2.5-opt local search~\cite{stattenberger2007neighborhood}. 
The resulting permutation $P^*$ is applied to all weight matrices within the clique. This promotes the weight smoothness and preserves the functionality of the network.

Since each individual mSHP problem is only correlated to one clip graph, we can solve the optimal $P$ in a relative small scale, very efficently. In fact, with $\leq 20$ cliques per network, the total computation time is $<4$ sec.


\noindent\textbf{Cross-Model Smoothness.} Another crucial challenge is to perverse the generalization behavior of the INR with different network configurations, which means, a small perturbation in the configuration, will eventually not affect the main model's performance. We address this by adding coordinate variation in the INR's learning process.

\setlength{\intextsep}{8pt}%
\setlength{\columnsep}{8pt}%
\begin{wrapfigure}{r}{0.52\textwidth}
    \includegraphics[width=\linewidth]{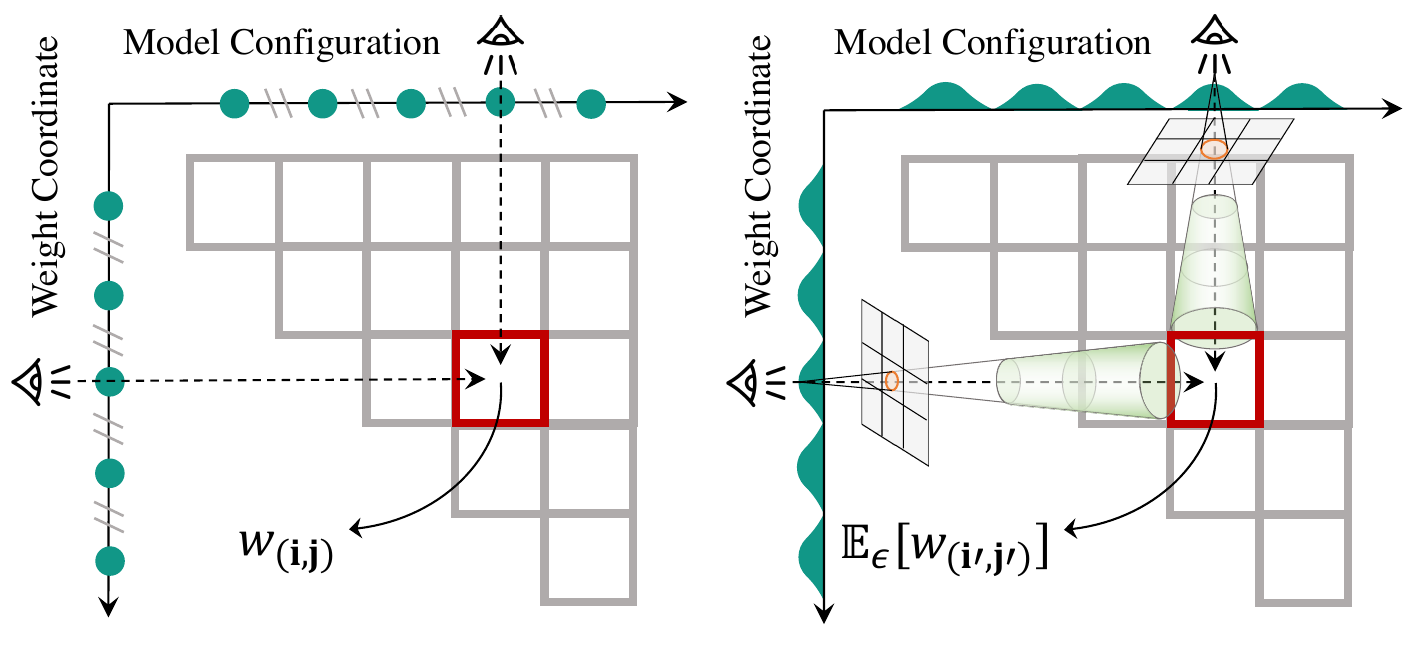}
    \caption{\textbf{Cross-model smoothness} via coordinate perturbation. Unlike the predict weights in discrete grid~\textbf{(Left)}, our INR  predicts weight as the expectation within a small neighborhood~\textbf{(Right)}. }
    \label{fig:sampling}
\end{wrapfigure}

During training, rather than using fixed coordinates and model sizes as in Equation~\ref{eq:1}, we introduce slight variations to the input coordinates. Specifically, we add a small perturbation $\bm\epsilon$ to the input coordinate $(\mathbf{i}',\mathbf{j}') = (\mathbf{i},\mathbf{j}) + \bm\epsilon$, where $\bm\epsilon$ is drawn uniformly from $\text{U}(-\mathbf{a}, \mathbf{a})$. This strategy aims to minimize the expected loss $\mathbb{E}_{\bm\epsilon \in \text{U}(-\mathbf{a}, \mathbf{a})}[\mathcal{L}]$.

For model evaluation, we sampling weight from a small neighborhood, as illustrated in Figure~\ref{fig:sampling}. We compute this by averaging the weights obtained from multiple input, each perturbed by different $\bm\epsilon \in \text{U}(-\mathbf{a}, \mathbf{a})$:
{\small\begin{equation}
\Bar{w}_{(\mathbf{i},\mathbf{j})}= \mathbb{E}_{\bm\epsilon \in \text{U}(-\mathbf{a}, \mathbf{a})}[w_{(\mathbf{i}',\mathbf{j}')}] \approx \frac{1}{K} \sum_{K} F(\gamma_{\text{PE}}(\mathbf{v}');\theta)
\end{equation}}
This is implemented by inferring the INR $K=50$ times with varied sampled inputs $\mathbf{v}'$ and then computing the average of these weights to parameterize the main network. This approach is designed to enhance the stability and reliability of the INR under different configurations.

\subsection{Training and Optimization}
\label{sec:train}
Our approach optimizes the INR, denoted as $F(\cdot;\theta)$, to accurately predict weights for the main network of different configurations. It pursues two primary goals: approximating the weights of the pretrained network 
 $f(\cdot;\mathbf{W}_{\text{original}})$, and minimizing task-specific loss across a range of randomly sampled networks. As such, the optimization leverages a composite loss function, divided into three distinct components: task-specific loss, reconstruction loss, and regularization loss.

\noindent\textbf{Task-specific Loss.} Denoted as \(\mathcal{L}_{\text{task}}(y, \hat{y}(\mathbf{W}))\), this measures the difference between actual labels $y$ and predictions $\hat{y}$, based on weights $\mathbf{W}$ from the INR.

\noindent\textbf{Reconstruction Loss.} This element, expressed as $\mathcal{L}_{\text{recon}} = ||\mathbf{W}_{\text{original}}||^2_{2}||\mathbf{W} - \mathbf{W}_{\text{original}}||^2$, assesses how close the INR-derived weights $\mathbf{W}$ to the ideal weights $\mathbf{W}_{\text{original}}$, weighted by the magnitude $||\mathbf{W}_{\text{original}}||^2_{2}$.

\noindent\textbf{Regularization Loss.} Symbolized as $\mathcal{L}_{\text{reg}} =  ||\mathbf{W}||^2$. This introduces L2 norm regularization on the predicted weights, to prevent overfitting by controlling the complexity of the derived model~\cite{krogh1991simple,loshchilov2017decoupled}.

We minimize the composite objective by sampling different points on the model space
{\small \begin{equation}
    \min_{\theta} \mathbb{E}_{\mathbf{i},\mathbf{j},\bm\epsilon}[\mathcal{L}] = \min_{\theta}\mathbb{E}_{\mathbf{i},\mathbf{j},\bm\epsilon}[\mathcal{L}_{\text{task}} + \lambda_1 \mathcal{L}_{\text{recon}} + \lambda_2 \mathcal{L}_{\text{reg}}] \label{eq:trainloss}
\end{equation}}

This loss function ensuring not only proficiency in the primary task through precise weight, but also bolstering model robustness via regularization. During training, we iteratively evaluates various combinations $(\mathbf{i},\mathbf{j})$, striving to minimize the expected loss. The loss function is backpropagated from the main network to the INR as follows:
{\small
\begin{equation}
    \nabla_{\theta} \mathcal{L} = \frac{\partial \mathcal{L}_{\text{task}}}{\partial W} \frac{\partial W}{\partial \theta} + \lambda_1 \frac{\partial \mathcal{L}_{\text{recon}}}{\partial \theta} + \lambda_2\frac{\partial \mathcal{L}_{\text{reg}}}{\partial \theta}
\end{equation}}
This equation represents the gradient of the loss with respect to $\theta$.

%% file: sec/5_exp.tex
\section{Experiments}
In this section, we present our experimental analysis and various applications of \texttt{NeuMeta}, spanning classification, semantic segmentation, and image generation. To substantiate our design choices, we conduct different ablation studies. Additionally, we delve into exploring the properties of the learned weight manifold.

\subsection{Experimental Setup}
\noindent\textbf{Datasets and Evaluation.} We evaluate the proposed method on 3 tasks across 6 different visual datasets. For image classification, we select 4 dataset: MNIST~\cite{lecun-mnisthandwrittendigit-2010}, CIFAR10, CIFAR100~\cite{Krizhevsky09learningmultiple} and ImageNet~\cite{deng2009imagenet}. Training includes horizontal flip augmentation, and we report top-1 accuracy. We also report the training time and the final size of stored parameters to evaluate our savings.

In semantic segmentation, we utilize PASCAL VOC2012~\cite{Everingham10}, a standard dataset for object segmentation tasks. We utilize its augmented training set~\cite{hariharan2011semantic}, incorporating preprocessing techniques like random resize crop, horizontal flip, color jitter, and Gaussian blur. Performance is quantified using mean Intersection-over-Union (mIOU) and F1 score, averaged across 21 classes.

For image generation, we employ MNIST and CelebA~\cite{liu2015faceattributes}. A vanilla variational auto-encoder (VAE) fits the training data, with evaluation based on reconstruction MSE and negative log-likelihood (NLL).

\noindent\textbf{Implementation Details.} Our INR utilizes MLPs with ReLU activation, comprising five layers with residual connections and 256 neurons each. We employ a block-based INR approach, where each parameter type, such as weights and biases, is represented by a separate MLP. The positional embedding frequency is set to 16. Optimization is done using Adam~\cite{KingBa15} with a 1e-3 initial learning rate and cosine decay, alongside a 0.995 exponential moving average. During training, we maintain the balance between different objectives with $\lambda_1 = 1$ and $\lambda_2 = 1e-4$. In each training batch, we sample one network configuration, update a random subset of layers in the main network, computing gradients for the INR, to speed up training. The configuration pool, created by varying the channel number of the original network, utilizes a \emph{compression rate} $ \gamma = 1-\frac{\text{sampled channel number}}{\text{full channel number}}$. We randomly sample a network width with compress rate $\gamma \in [0,0.5]$ for training. For example, a 128-channel layer will have its width sampled from $64\sim128$.

For classification tasks, we apply LeNet~\cite{lecun-gradientbased-learning-applied-1998} on MNIST, ResNet20~\cite{he2016deep} on CIFAR10 and CIFAR100, and ResNet18 and ResNet50 on ImageNet, using batch sizes of 128 (MNIST, CIFAR) and 512 (ImageNet). We the INR train for 200 epochs. For segmentation task, we use the U-Net~\cite{ronneberger2015u} with the ResNet18 backbone.  The details are mentioned in the supplementary material.

\noindent\textbf{Baselines.} Our \texttt{NeuMeta} model is benchmarked against three family of methods: Structure Pruning, Flexible Models, and Continuous-Width Models. 
\begin{itemize}
    \item \textbf{Individually Trained Model.} Each models is trained separately.
    \item \textbf{Structure Pruning.} We evaluate against pruning methods that eliminate channels based on different criteria. This includes Weight-based pruning (removing neurons with low $\mathcal{\ell}_1$/$\mathcal{\ell}_2$-norm), Taylor-based method (using gradients related to output), Hessian-based pruning (using Hessian trace for sensitivity), and Random pruning (random channel removal).
    \item \textbf{Flexible Model.}  We compare with the Slimmable network~\cite{DBLP:conf/iclr/YuYXYH19}, which trains subnetworks of various sizes within the main model for dynamic test-time resizing. We train the model on $\{0\%,25\%,50\%\}$ compressed ratio, and also test on 75\% compressed setup.
    \item \textbf{Continuous-Width NN.} Comparison is made with the Integral Neural Network~\cite{Solodskikh_2023_CVPR}, which uses kernel representation for continuous weight modeling. For comparison, we focus on the uniform sampling method, as its learned sampling technique does not support resizing. We train the model on $[0\%,50\%]$ compress rate range, but also test on other values, like 75\%.
\end{itemize}
We ensure that all compression is applied uniformly for all methods, guaranteeing that all models compared have the exactly same cost for inference.

\subsection{Enhancing Efficiency by Morphing the Networks}
\begin{figure}[t]
    \centering
    \begin{subfigure}{0.24\linewidth}
        \includegraphics[width=\linewidth]{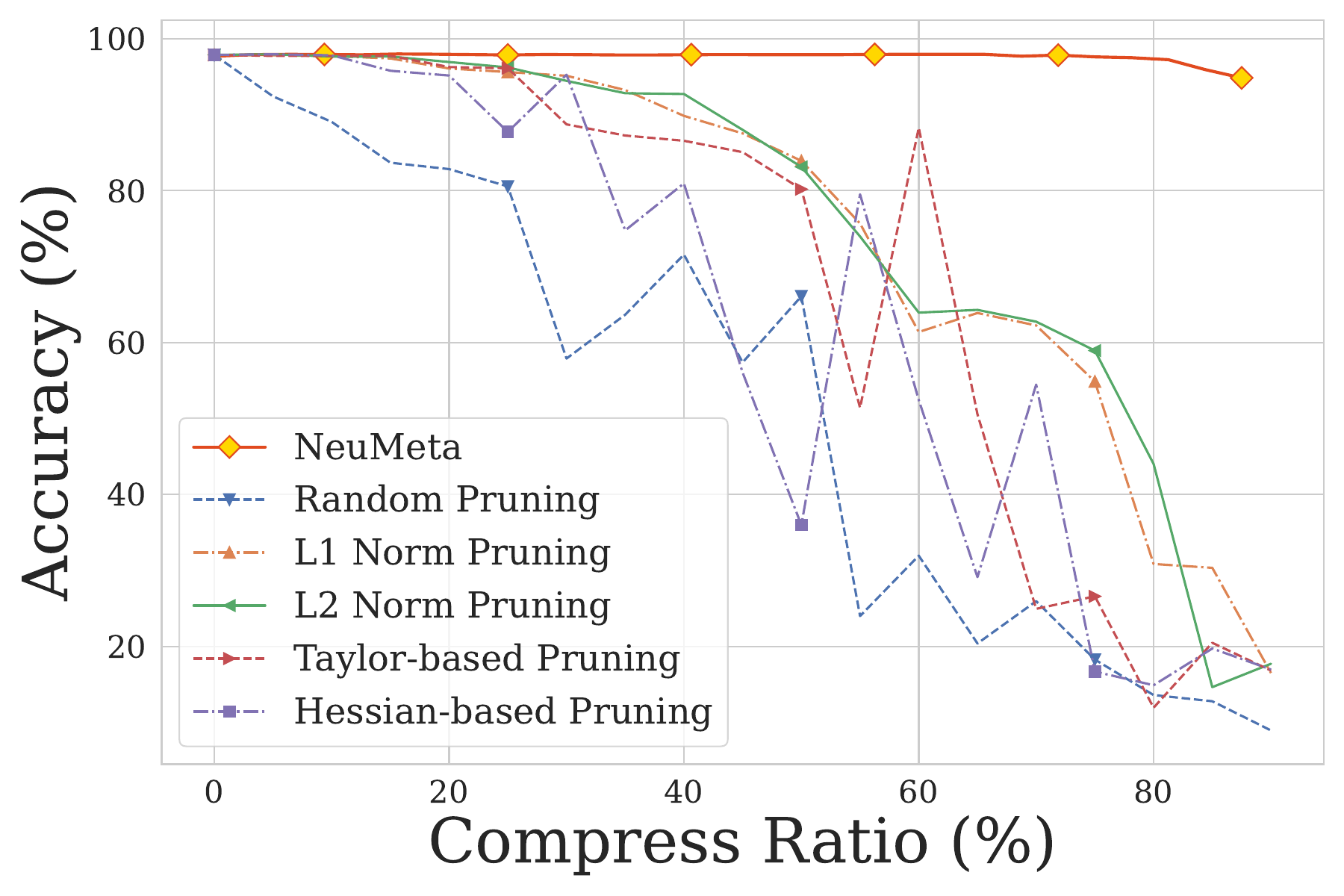} 
        \caption{LeNet on MNIST}
        \label{fig:mnist}
    \end{subfigure}
    \hfill 
    \begin{subfigure}{0.24\linewidth}
        \includegraphics[width=\linewidth]{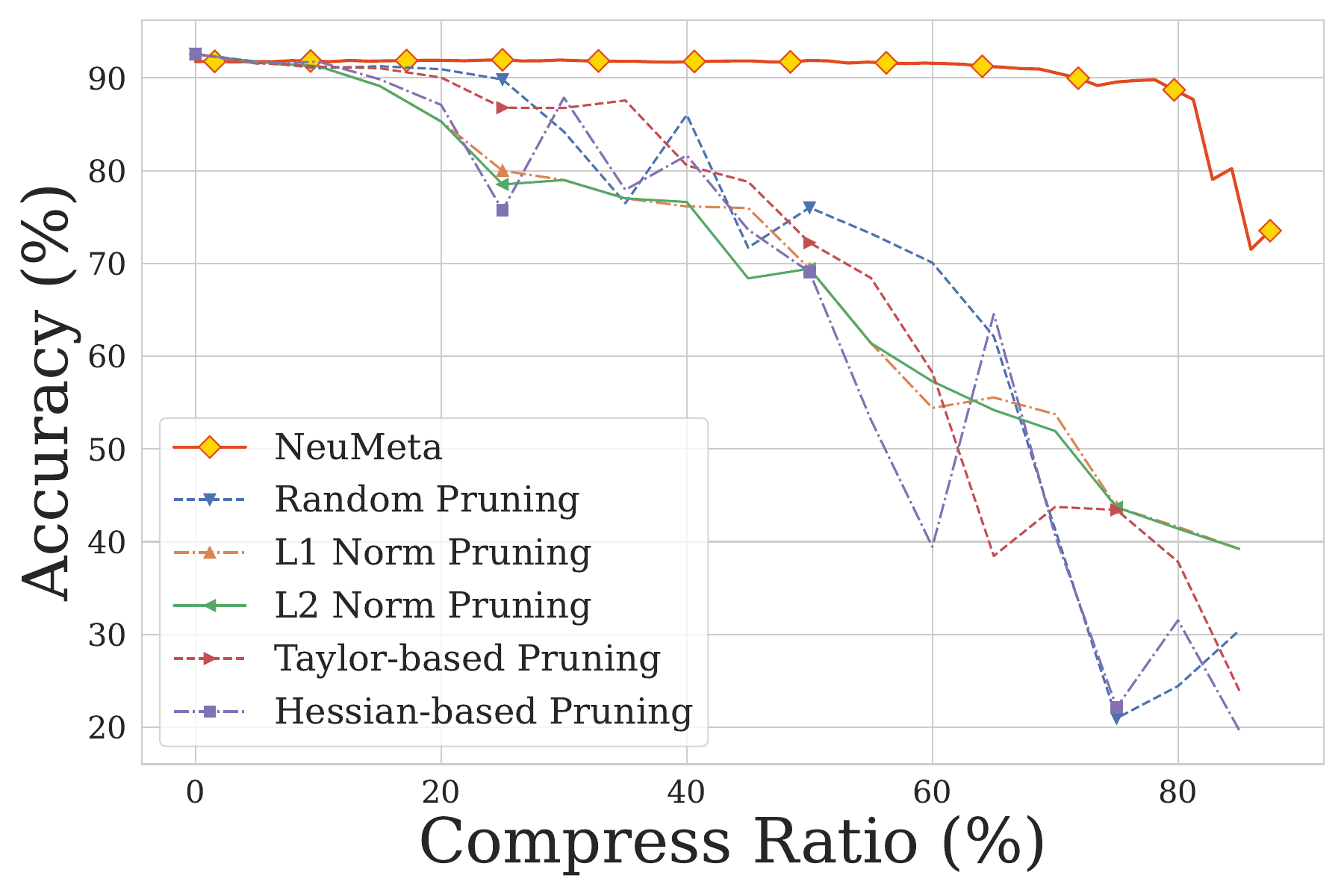} 
        \caption{R20 on CIFAR10}
        \label{fig:cifar10}
    \end{subfigure}
\hfill
    \begin{subfigure}{0.24\linewidth}
        \includegraphics[width=\linewidth]{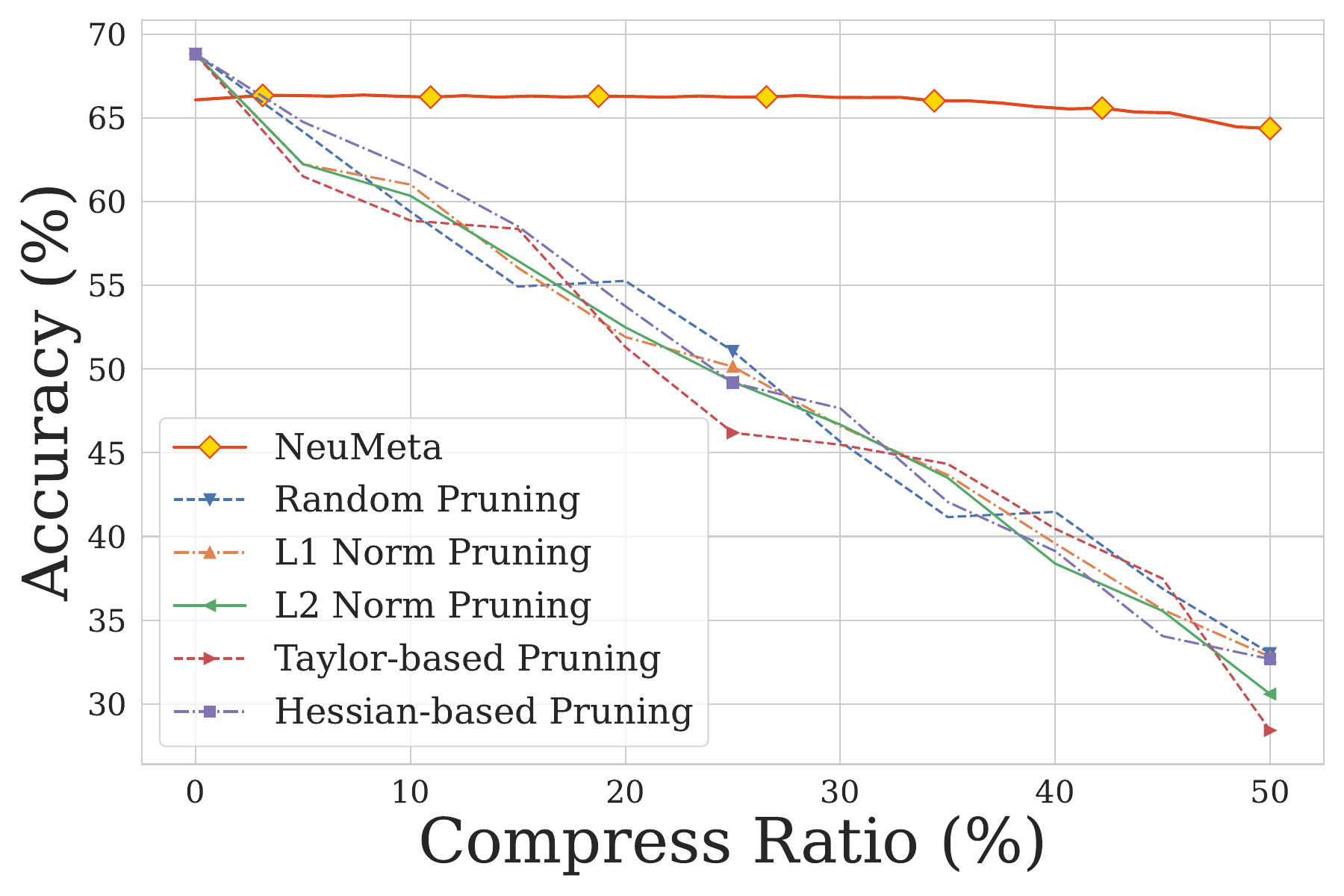} 
        \caption{R20 on CIFAR100}
        \label{fig:cifar100}
    \end{subfigure}
        \hfill
    \begin{subfigure}{0.24\linewidth}
        \includegraphics[width=\linewidth]{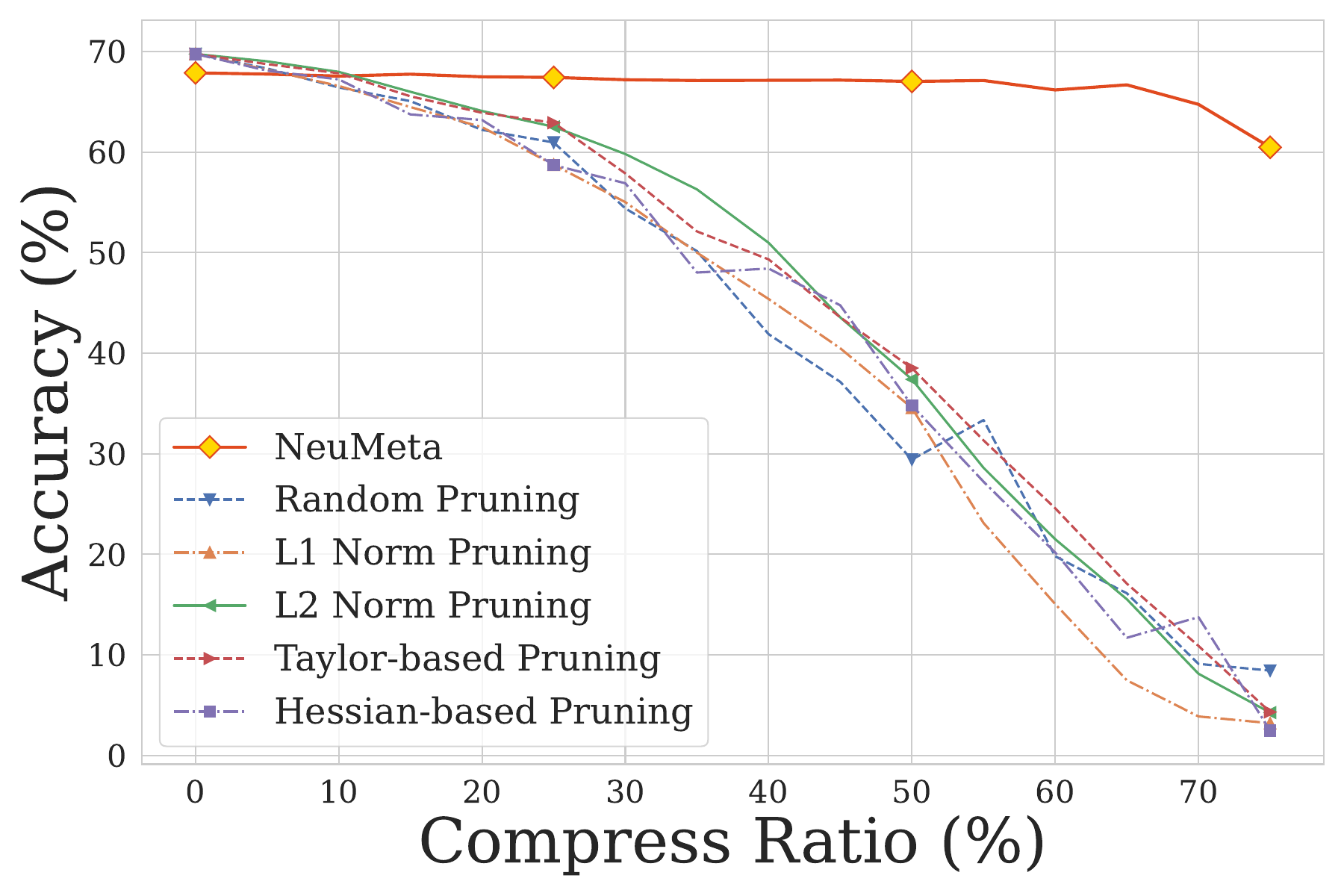} 
        \caption{R18 on ImageNet}
        \label{fig:imagenet}
    \end{subfigure}
    \caption{Accuracy comparison of \texttt{NeuMeta} versus different structure pruning methods on MNIST, CIFAR10, CIFAR100 and ImageNet. Our method consistently outperforms pruning-based methods. R18 and R20 are short for ResNet18 and ResNet20.}
    \label{fig:prune} 
\end{figure}

\noindent\textbf{Image Classification.} As depicted in Figure~\ref{fig:prune}, in the realm of image classification, \texttt{NeuMeta} consistently surpasses existing pruning-based methods in accuracy across MNIST, CIFAR10, CIFAR100, and ImageNet datasets at various compression ratios. It is worth-noting that, pruning-based methods show a marked accuracy decrease, approximately 5\% on ImageNet and 6\% on CIFAR100, when the compression ratio exceeds 20\%. Conversely, \texttt{NeuMeta} retains stable performance up to 40\% compression. However, a minor performance reduction is noted in our full-sized model, highlighting a limitation in the INR's ability to accurately recreate the complex pattern of network weights.

Table~\ref{tab:unified_compression_performance} compares \texttt{NeuMeta} with Slimable NN and INN, including \emph{Oracle} results of independently trained models for reference. We stick to the same model size for all method, to ensure the comparision is fair. Remarkably, \texttt{NeuMeta} often surpasses even these oracle models on large compress rate. This success is attributed to the preserved smoothness across networks of varying sizes, which inadvertently enhances smaller networks. Our approach outperforms both Slimable NN and the kernel representation in INN. Notably, at an untrained compression ratio of 75\%$^\dagger$, other methods significantly underperform. 

Furthermore, when evaluating total training time and parameter storage requirements, our approach demonstrates improved efficiency. Unlike the exhaustive individual model training and storage approach, other methods achieve some level of savings. However, Slimable NN's separate storage for BN parameters still renders it less efficient. Our method achieves the least storage size by storing a few MLPs instead of the original parameters, thus reducing the overall parameter count even below that of a single model.

                                          


\begin{table}[t]
\setlength\tabcolsep{1pt}
\scriptsize


\centering
\begin{tabular}{l|c|c|c|c|c|c}
\toprule
& \multicolumn{6}{c}{ResNet20 on CIFAR10}  \\
\cline{1-7}
\multirow{2}{*}{Method} & $\gamma=0\%$ & $\gamma=25\%$ & $\gamma=50\%$ & $\gamma=75\%$$^\dagger$ & \multirow{2}{*}{\makecell{Total Train Cost \\ (GPU hours)}} & \multirow{2}{*}{\makecell{Stored \\ Params}} \\
\cline{2-5}
& Acc & Acc  & Acc  & Acc  & & \\ 
\midrule
\rowcolor{gray!10} Individual & 92.60 &  90.65 & 89.57 &  87.04 & 5.3 & 0.67M\\
Slimable~\cite{DBLP:conf/iclr/YuYXYH19} & 90.44 & 90.44  & 88.41 & \textcolor{gray}{18.56}  & 1.6 & 0.35M\\
INN~\cite{Solodskikh_2023_CVPR} & 91.33 &  90.50 & 89.24  & \textcolor{gray}{71.70} & 1.8 & 0.27M \\\hline
\rowcolor{cyan!10} Ours & \textbf{91.76}  & \textbf{91.32} & \textbf{90.56}  & \textbf{89.56} &  \textbf{1.3} & \textbf{0.20M} \\
\midrule
& \multicolumn{6}{c}{ResNet20 on CIFAR100}  \\
\cline{1-7}
\multirow{2}{*}{Method} &$\gamma=0\%$ & $\gamma=25\%$ & $\gamma=50\%$ & $\gamma=75\%$$^\dagger$ & \multirow{2}{*}{\makecell{Total Train Cost \\ (GPU hours)}} & \multirow{2}{*}{\makecell{Stored \\ Params}} \\
\cline{2-5}
& Acc & Acc & Acc & Acc  & & \\ 
\midrule
\rowcolor{gray!10} Individual & 68.83 & 66.37 & 64.87 & 61.37 & 5.5 & 0.70M \\
Slimable~\cite{DBLP:conf/iclr/YuYXYH19} &64.44   & 64.01&63.38& \textcolor{gray}{1.59} & 1.5 & 0.37M\\
INN~\cite{Solodskikh_2023_CVPR} & 65.86& 65.53 & 63.35 & \textcolor{gray}{27.60} & 1.9 & 0.28M \\\hline
\rowcolor{cyan!10} Ours & \textbf{66.07}  & \textbf{66.23} & \textbf{65.36} & \textbf{62.62} & \textbf{1.4} & \textbf{0.20M}  \\
\bottomrule
\end{tabular}
\caption{Accuracy comparison of ResNet20 on CIFAR10 and CIFAR100 at different compression ratios. $^\dagger$ The 75\% compression ratio wasn't applied in training. }
\label{tab:unified_compression_performance}
\end{table}

\begin{figure}[t]
    \centering
    \begin{subfigure}{0.49\linewidth}
       \includegraphics[width=\textwidth]{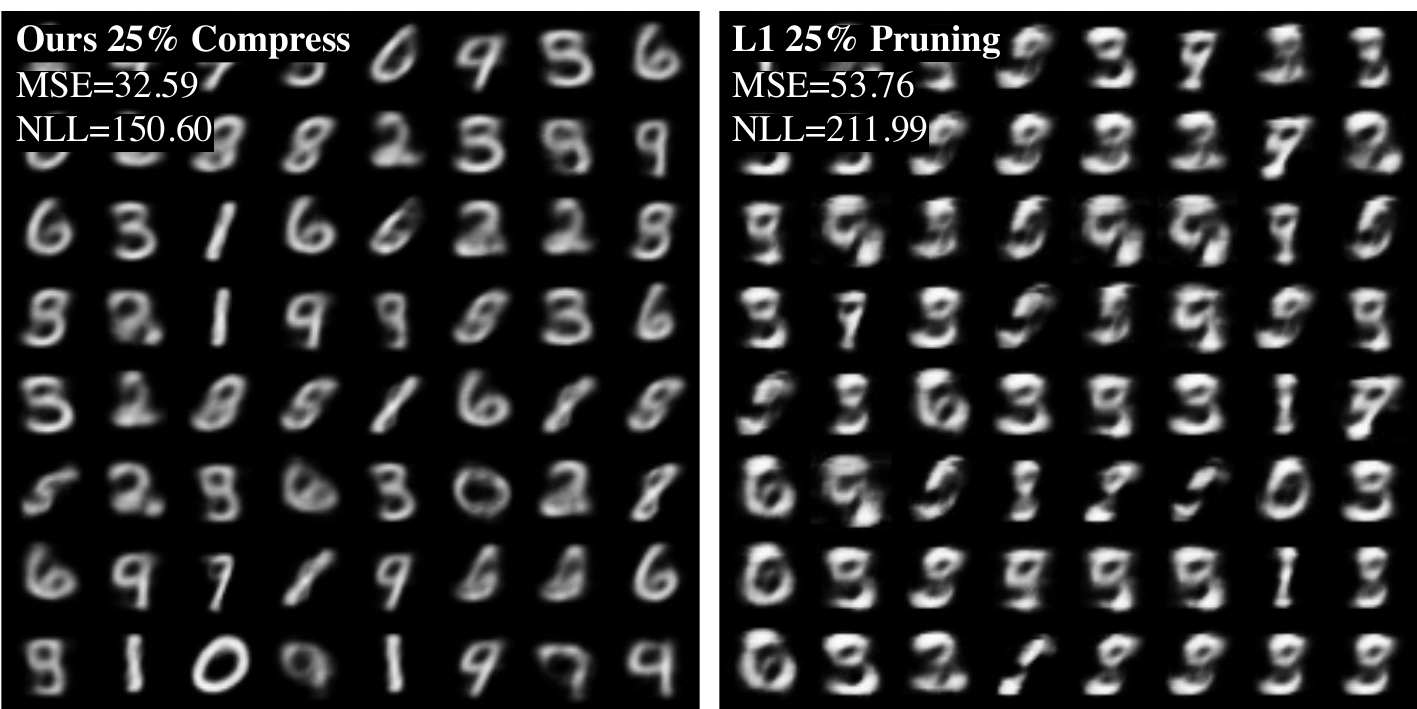}
        \caption{Results on MNIST}
        \label{fig:visualize_vae_mnist}
    \end{subfigure}
    \hfill 
    \begin{subfigure}{0.49\linewidth}
         \includegraphics[width=\textwidth]{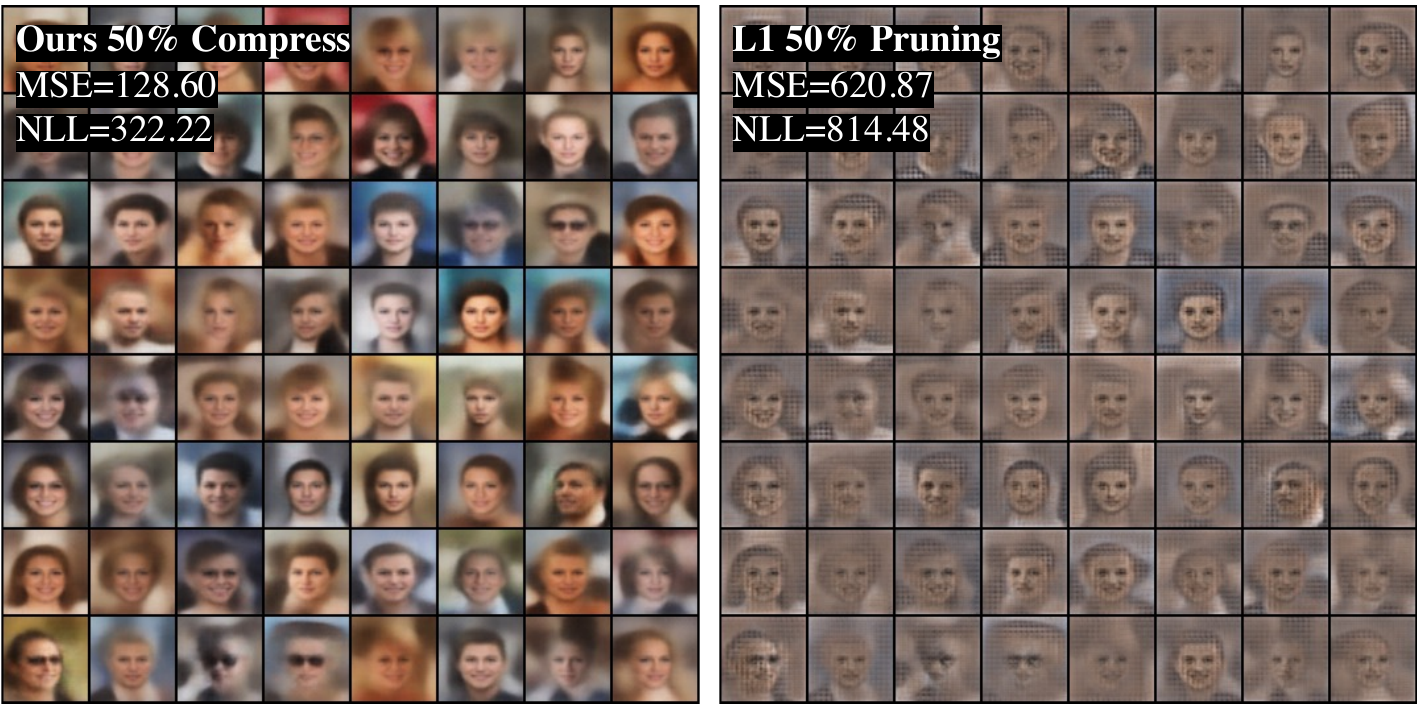}
        \caption{Results on CelebA}
        \label{fig:visualize_vae_celeba}
    \end{subfigure}
    \caption{VAE Visualizations on MNIST and CelebA Datasets on the same compress rate. Lower NLL and MSE indicates better performance.}
    \label{fig:vae_visualizations} 
\end{figure}

\begin{figure}[t]
\centering
\begin{minipage}{.49\textwidth}
  \centering
    \begin{subfigure}[b]{0.48\linewidth}
        \includegraphics[width=\textwidth]{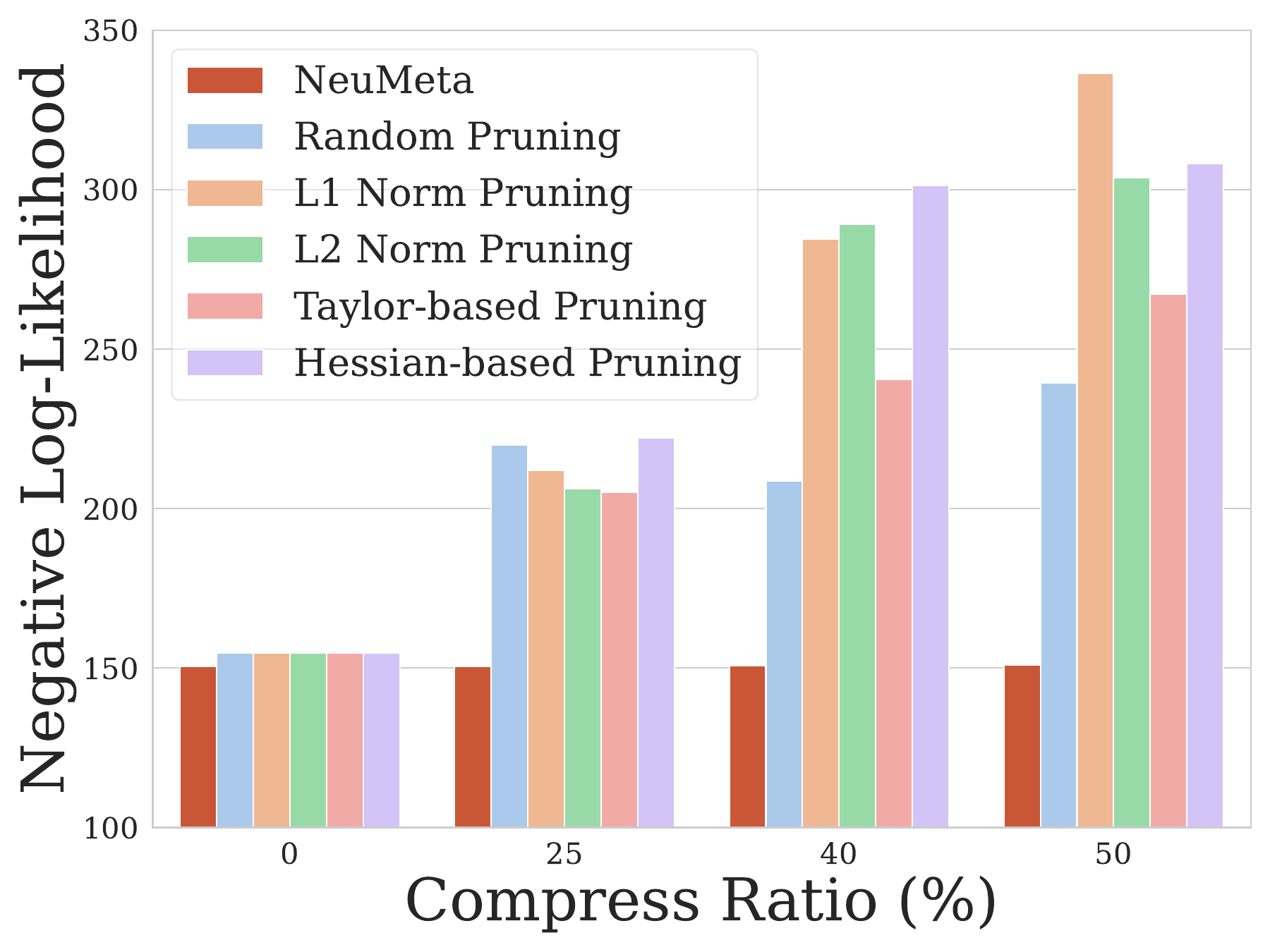}
        \caption{MNIST Compress Rate \textit{vs.} NLL}
        \label{fig:mnist_vae}
    \end{subfigure}%
    \hfill 
    \begin{subfigure}[b]{0.48\linewidth}
        \includegraphics[width=\textwidth]{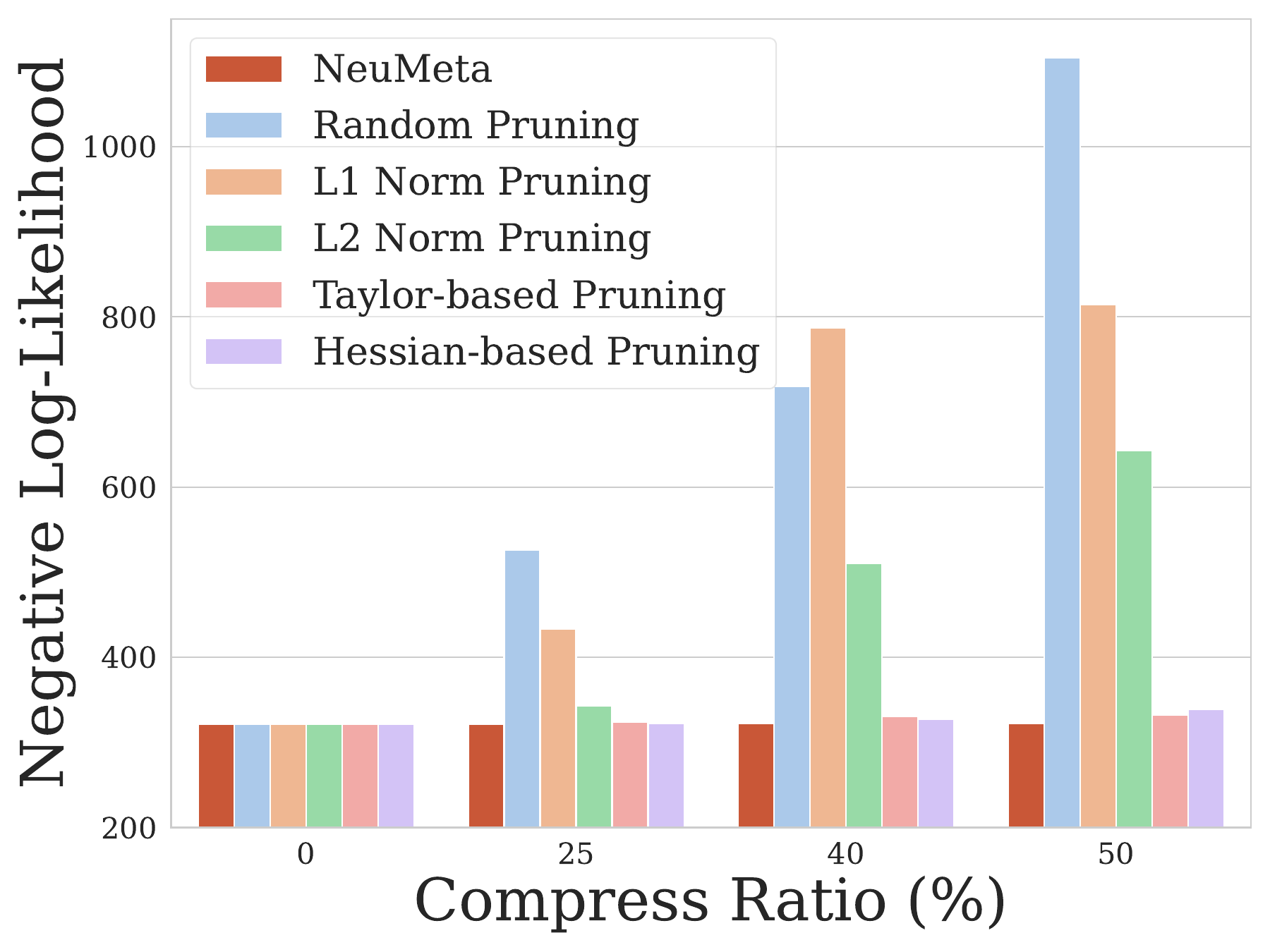}
        \caption{CelebA Compress Rate \textit{vs.} NLL}
        \label{fig:celeba_vae}
    \end{subfigure}
    \caption{Comparative analysis of compress rate and NLL on different datasets. Lower NLL indicates better performance.}
    \label{fig:ratio_nll_comparison}
\end{minipage}%
\hfill
\begin{minipage}{.49\textwidth}
\renewcommand{\arraystretch}{2}
\setlength\tabcolsep{1pt}
\tiny
\centering
\begin{tabular}{l|c|c|c|c|c|c}
\toprule

 \multirow{2}{*}{Method}& \multicolumn{2}{c|}{25\%} & \multicolumn{2}{c|}{50\%} & \multicolumn{2}{c}{75\%$^\dagger$} \\ \cline{2-7} 
   & mIOU & F1 & mIOU & F1 & mIOU & F1 \\ \midrule
\rowcolor{gray!10}Individual              &84.70 & 90.63 & 83.14         & 89.59       & 82.79         & 89.36       \\ 

Slimmable~\cite{DBLP:conf/iclr/YuYXYH19}&81.09 & 88.14 & 80.92 & 88.03 & \textcolor{gray}{61.19} & \textcolor{gray}{72.78}      \\ \hline
\rowcolor{cyan!10}Ours                   & \textbf{81.94} & \textbf{88.75}  & \textbf{81.93}         & \textbf{88.74}       & \textbf{81.94}        & \textbf{88.75}       \\ \bottomrule
\end{tabular}
\caption{Comparison of different methods across compressed ratio for U-Net. $^\dagger$ The 75\% compression ratio wasn't seen in training.}
\label{tab:seg_comparison}
\end{minipage}
\end{figure}

\noindent\textbf{Semantic Segmentation.} For semantic segmentation on the PASCAL VOC2012 dataset, \texttt{NeuMeta} demonstrates superior performance in Table~\ref{tab:seg_comparison}. It surpasses the Slimmable network that requires hard parameter sharing, especially at an untrained 75\% compression rate. On this setup, we show a significant improvement of 20 mIOU. However, for complex tasks like segmentation, a slight performance drop is observed at smaller compression rate. It is attributed to the INR's limited representation ability. More results is provided in the supplementary.

\noindent\textbf{Image Generation.} We implement \texttt{NeuMeta} to generate images on MNIST and CelebA, using VAE. Since Slimable NN and INN  haven't been previously adapted for VAE before, we only compare with the pruning method, in Figure~\ref{fig:ratio_nll_comparison}.  Our approach demonstrated superior performance in terms of lower negative log-likelihood (NLL) across various compression ratios. For example, we visualize the generated results of when compressed by 25\% for MNIST and 50\% for CelebA in Figure~\ref{fig:vae_visualizations}. Compared with the $\mathcal{\ell}_1$-based pruning, our method significantly improved reconstruction MSE from 53.76$\to$32.58 for MNIST and from 620.87$\to$128.60 for CelebA. Correspondingly, the NLL was reduced by 61.33 for MNIST and 492.26 for CelebA.


\subsection{Exploring the Properties for NeuMeta}
As we represent the weights as a smooth manifold, we investigate its effects on network. Specifically, we compare \texttt{NeuMeta} induced networks, with individually trained models and models distilled~\cite{hinton2015distilling} from full-sized versions.

\noindent\textbf{NeuMeta promote feature similarity.} We analyzed the last layer features of ResNet20 trained on CIFAR10, particularly from \texttt{layer3.2}, using linear central kernel alignment~(CKA)~\cite{kornblith2019similarity} score between each resized and the full-sized model. The result is shown in Figure~\ref{fig:similarity}~(Top). It reveals higher feature map correlations across models compared to other methods, indicating that \texttt{NeuMeta} encourages similar network representations across different sizes.

\noindent\textbf{NeuMeta as Implicit Knowledge Distillation.} We also report the the pair-wise output KL divergence in Figure~\ref{fig:similarity}~(Bottom), a key metric in knowledge distillation~\cite{hinton2015distilling}. Individually trained models show higher divergence, whereas both KD and \texttt{NeuMeta} result in reduced divergence. These results imply that \texttt{NeuMeta} not only aligns internal representations but also ensures consistent network outputs, as an implicit form of  distillation.

\subsection{Ablation Study}

\begin{figure}[t]
\centering
\begin{minipage}{.48\textwidth}
  \centering
    \includegraphics[width=\linewidth]{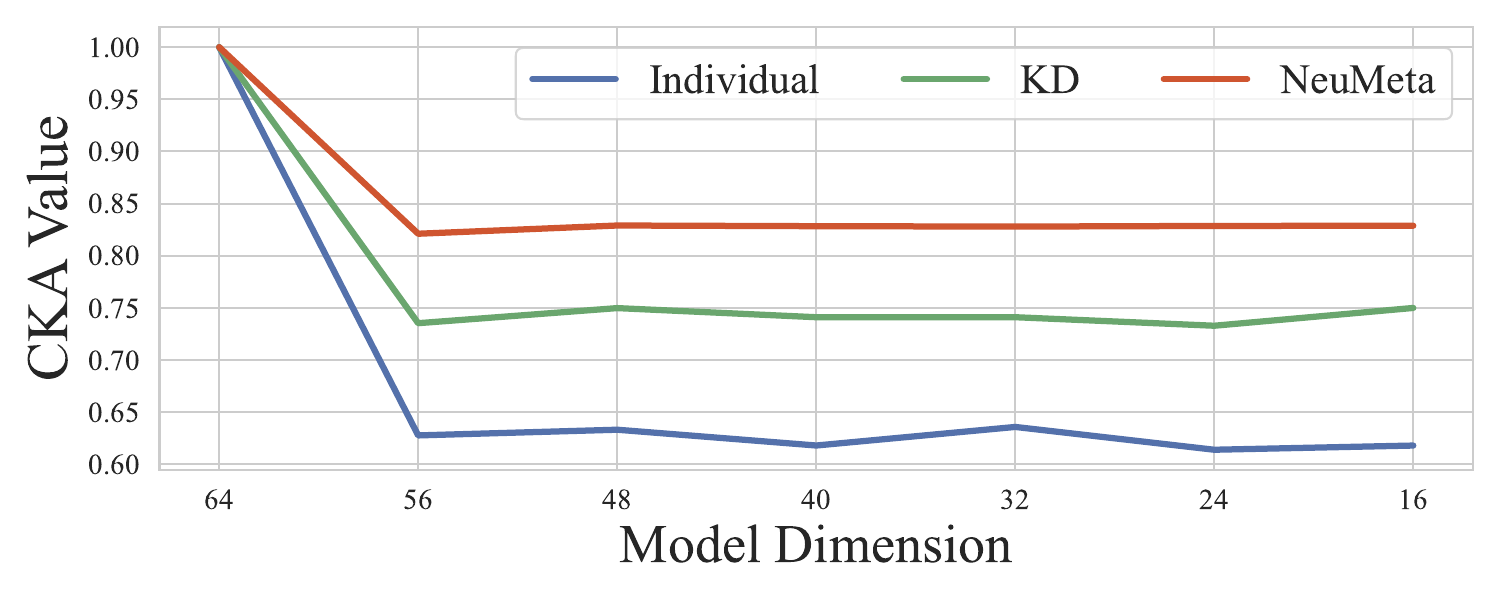}
    \includegraphics[width=\linewidth]{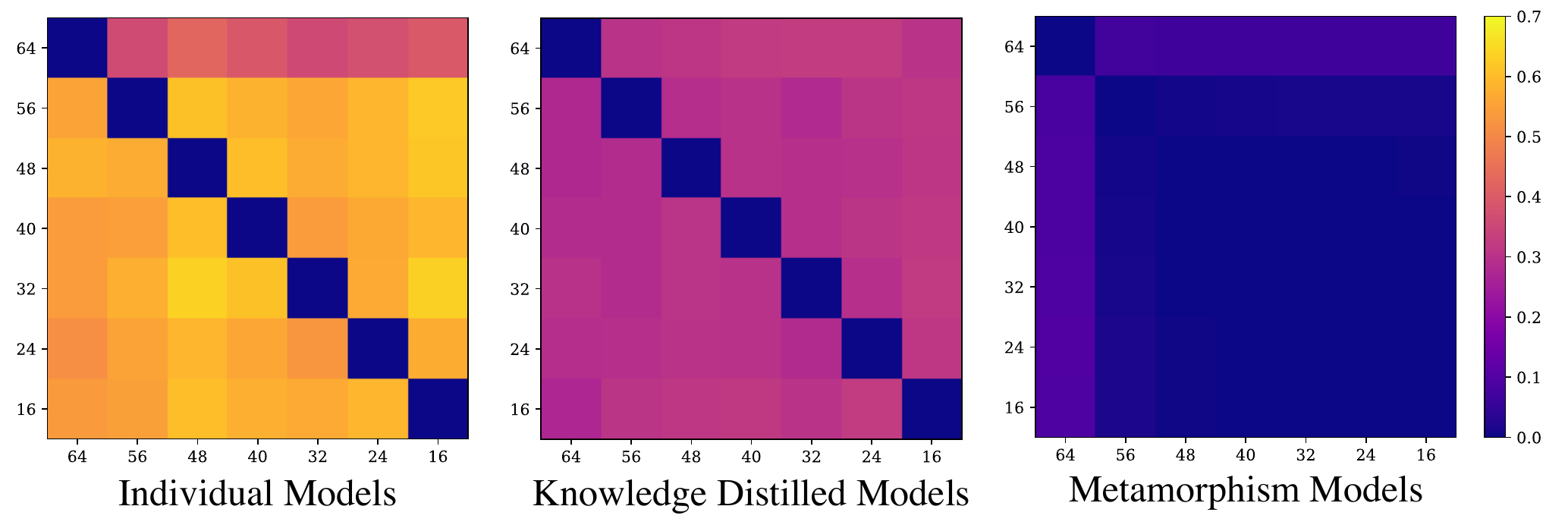}
    \caption{Similarity Analysis Between Models. \textbf{(Top)} the CKA comparison between the full model and various other models of different sizes. \textbf{(Bottom)} heatmap of the output KL divergence for each pair of models.}
    \label{fig:similarity}
\end{minipage}%
\hfill
\begin{minipage}{.48\textwidth}
\renewcommand{\arraystretch}{0.9}
  \centering
    \includegraphics[width=\linewidth]{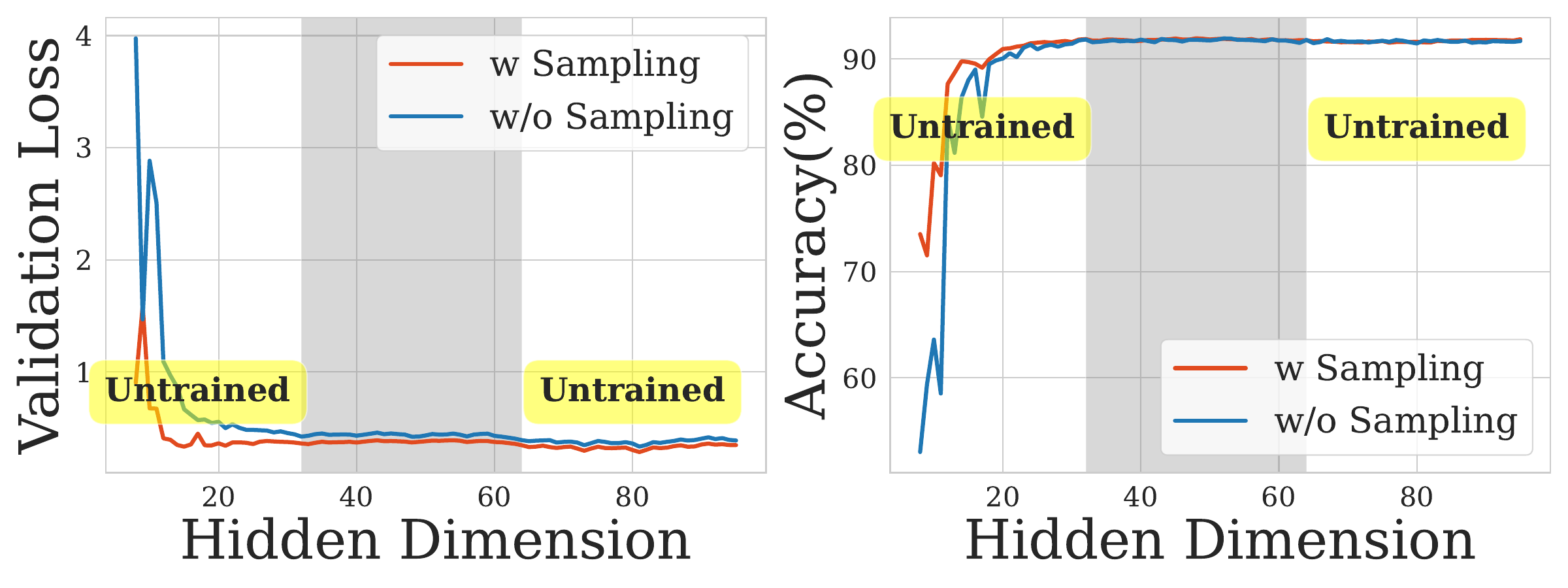}
    \caption{Ablation study with or without manifold sampling.}
    \label{fig:manifold sampling}
    \scriptsize
    \centering
    \begin{tabular}{l|ccc|c}
    \toprule
       No. & Weight Permutation  & $\lambda_1$ & $\lambda_2$ & \textbf{Accuracy} \\
        \midrule
     1 & \xmark &0& 1e-4 & 73.56\\
       2 & \xmark &1& 1e-4 & 80.33 \\
       3  & \cmark  &1& 0 & 64.37\\
      4& \cmark&1& 1e-4 & \textbf{91.84}\\
     5  &  \cmark &10& 1e-4 & 91.73\\
       6 & \cmark  &100& 1e-4 & 91.47\\
        \bottomrule
    \end{tabular}
    \caption{Ablation study for weight permutation and objective hyperprameters on CIFAR10 ResNet20.}
    \label{tab:ablation}
\end{minipage}
\end{figure}

\textbf{Weight Permutation.} To validate the effectiveness of our permutation strategy, we analyzed its impact on CIFAR10 accuracy. The comparison of Exp~2 and 4 in Table~\ref{tab:ablation} demonstrates a significant 11.51 accuracy increase due to our permutation strategy. Detailed comparisons of our mSHP-based method with the TSP solution from~\cite{Solodskikh_2023_CVPR} are presented in the supplementary material. It shows that our mSHP-based solution achieved lower weight total variation score, indicating superior with-in model smoothness.

\noindent\textbf{Objective.} We verify different terms in our training objective in Eq~\ref{eq:trainloss}. From Exp 1, 4-6 in Table~\ref{tab:ablation}, we find the optimal reconstruction weight $\lambda_1=1$ yields the best performance. Comparing Exp 3 and 4, we observe a performance boost with a weight penalty term at $\lambda_2 =1e-4$.

\noindent\textbf{Manifold Sampling.} Figure~\ref{fig:manifold sampling} evaluates our manifold sampling method with ResNet20 on CIFAR10. Sampling from the weight manifold neighborhood consistently improves performance, especially in untrained model sizes.

%% file: sec/6_conclusion.tex
\section{Conclusion}
This paper presents Neural Metamorphosis (\texttt{NeuMeta}), a novel paradigm that builds self-morphable neural networks. Through the training of neural implicit functions to fit the continuous weight manifold, \texttt{NeuMeta} can dynamically generate tailored network weights, adaptable across a variety of sizes and configurations. A core focus of our approach is to maintain the smoothness of weight manifold, enhancing the model's fitting ability and adaptability to novel setups. Experiments on image classification, generation and segmentation indicate that, our method maintain robust performance, even under large compression rate. 